\documentclass[10pt,twocolumn,letterpaper]{article}

\usepackage[pagenumbers]{cvpr} % To force page numbers, e.g. for an arXiv version

\usepackage{amsmath}
\usepackage{amssymb}
\usepackage{mathrsfs}
\usepackage{multirow}
\usepackage{pifont}
\usepackage{color}
\usepackage{algorithm}

\usepackage{graphicx}
\usepackage{array}

\definecolor{cvprblue}{rgb}{0.21,0.49,0.74}
\usepackage[pagebackref,breaklinks,colorlinks,allcolors=cvprblue]{hyperref}

\def\paperID{8118} % *** Enter the Paper ID here
\def\confName{CVPR}
\def\confYear{2025}

\title{ConceptMaster: Multi-Concept Video Customization on Diffusion Transformer Models Without Test-Time Tuning}

\author{
  Yuzhou Huang$^{1,2,3{*}}$ 
  \hspace{9pt} Ziyang Yuan$^{2,4{*}}$ 
  \hspace{9pt} Quande Liu$^{2}$$^\dagger$ 
  \hspace{9pt} Qiulin Wang$^{2}$ \\
  \hspace{9pt} Xintao Wang$^{2}$
  \hspace{9pt} Ruimao Zhang$^{1}$$^\dagger$
  \hspace{9pt} Pengfei Wan$^{2}$ 
  \hspace{9pt} Di Zhang$^{2}$ 
  \hspace{9pt} Kun Gai$^{2}$ \\
  \vspace{-0.05cm}
\small$^1$Sun Yat-sen University \hspace{5pt}
\small$^2$Kuaishou Technology\hspace{5pt}
\small$^3$The Chinese University of Hong Kong, Shenzhen \hspace{5pt}
\small$^4$Tsinghua University\hspace{5pt}
\\
\small \url{https://yuzhou914.github.io/ConceptMaster/}
}

\begin{document}

\twocolumn[{
\renewcommand\twocolumn[1][]{#1}
\maketitle
\vspace{-12mm}
\begin{center}
    \centering
    \captionsetup{type=figure}
    \hspace*{-4mm}\includegraphics[width=1.04\linewidth]{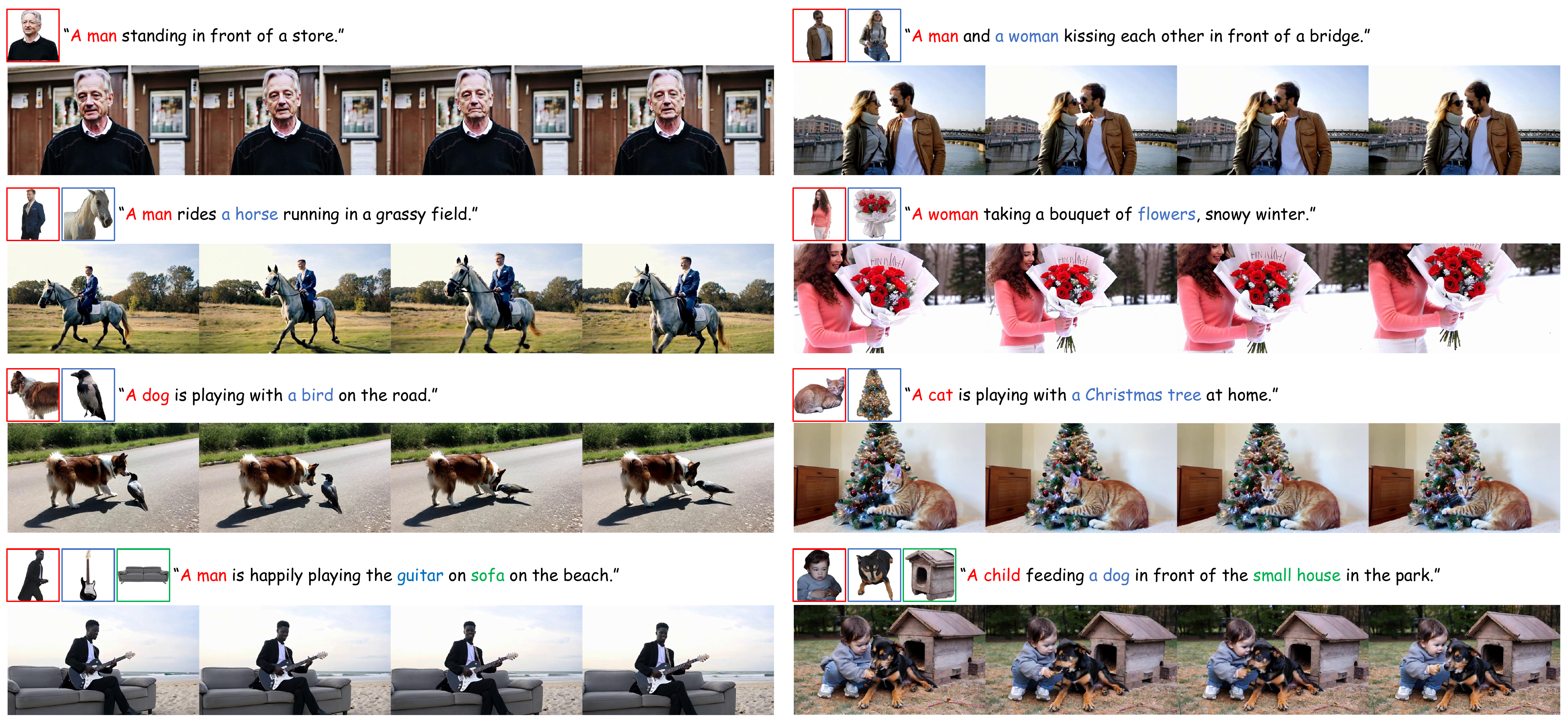}
    \vspace{-8mm}
    \captionof{figure}{We propose ConceptMaster, a Multi-Concept Video Customization (MCVC) method that can create high-quality concept-consistent videos based on given multiple reference images without test-time tuning. Representatively, we demonstrate ConceptMaster's video customization capacity on six scenarios, including \textit{1) multiple persons, 2) persons with livings, 3) persons with stuffs, 4) multiple livings, 5) livings with stuffs and 6) persons with both livings and stuffs.}}
    \label{fig:teaser}
    \vspace{-3mm}
\end{center}
}]

% 3. footnote
\maketitle
\let\thefootnote\relax\footnotetext{\begin{minipage}[t]{\textwidth}
$^*$ Work done during an internship at KwaiVGI, Kuaishou Technology \\
$^\dagger$ Corresponding author\hspace{3pt}
\end{minipage}}

\maketitle
\begin{abstract}
\vspace{-8mm}

% 文生视频发展，但是MCVC任务尚未解决
Text-to-video generation has made remarkable advancements through diffusion models. However, Multi-Concept Video Customization (MCVC) remains a significant challenge. 
% 我们认为有两大挑战
We identify two key challenges for this task: 1) the identity decoupling issue, where directly adopting existing customization methods inevitably mix identity attributes when handling multiple concepts simultaneously, and 2) the scarcity of high-quality video-entity pairs, which is crucial for training a model that can well represent and decouple various customized concepts in video generation.
% 因此我们提出ConceptMaster
To address these challenges, we introduce ConceptMaster, a novel framework that effectively addresses the identity decoupling issues while maintaining concept fidelity in video customization. 
% 我们怎么做的
% \todo{ConceptMaster employs a novel decoupled concept-embedding injection strategy, effectively separating and preserving the semantic uniqueness of individual concept embeddings. Complemented by a concept injector, our model can precisely query embeddings for each concept, even for highly similar visual concepts.}
% 这里给核心insight起了一个名字
Specifically, we propose to learn decoupled multi-concept embeddings and  inject them into diffusion models in a standalone manner, which effectively guarantees the quality of customized videos with multiple identities, even for highly similar visual concepts.
% 数据收集管线
To overcome the scarcity of high-quality MCVC data, we establish a data construction pipeline, which enables collection of high-quality multi-concept video-entity data pairs across diverse scenarios. 
% MC-Bench在六种场景下的三维度评估
A multi-concept video evaluation set is further devised to comprehensively validate our method from three dimensions, including concept fidelity, identity decoupling ability, and video generation quality, across six different concept composition scenarios. 
% 我们的方法更优 quantitative and qualitative
Extensive experiments demonstrate that ConceptMaster significantly outperforms previous methods for video customization tasks, showing great potential to generate personalized and semantically accurate  content for video diffusion models.

\end{abstract}
    
\section{Introduction}
\label{sec:intro}

% 1. 文生视频模型带动了视频定制化生成
Diffusion-based text-to-video generation models, trained on extensive text-video data pairs, have demonstrated remarkable success in generating high-quality videos from textual inputs~\cite{guo2023animatediff, chen2024videocrafter2, wang2023modelscope, videoworldsimulators2024, wu2022tune, villegas2022phenaki, singer2022make, luo2023videofusion, khachatryan2023text2video, he2022latent, esser2023structure, blattmann2023align}. 
% 出现了很多视频定制化生成工作
These advancements have sparked increasing interest in personalizing video generation through user-defined concepts. Recently, some methods have been proposed to produce customized videos using additional image guidance, with proven effectiveness for the customization on object~\cite{jiang2024videobooth}, human~\cite{he2024id}, style~\cite{liu2023stylecrafter}, etc.

% 2. tuning-based/pretrain-based 定制化生成，不区分图像和视频
% tuning-based，核心是每个新的概念都要优化后才能inference，且需要使用者提供few-shot examples，所以耗时且费力，无法实际应用
Existing approaches for concept customization primarily fall into two methodological categories: tuning-based solutions and pre-training-based methods. The tuning-based solutions~\cite{ruiz2023dreambooth, gal2022image, kumari2023multi, han2023svdiff, choi2023custom} typically first optimize model parameters (e.g., variants of LoRAs~\cite{hu2021lora} or fully training latent diffusion models~\cite{ruiz2023dreambooth}) each time for customized concepts and then incorporate them for inference.
% 重点是 tuning-based 有很大缺陷
However, these methods are computationally time-consuming and often require the manual collection of multiple reference samples, rendering them impractical in most time-sensitive and user-friendly scenarios.
% pretrain-based是数据驱动
Conversely, pre-training-based methods~\cite{wei2023elite, gal2023encoder, li2024blip, chen2024anydoor, ye2023ip, li2024photomaker, wang2024instantid, wang2024ms, jiang2024videobooth, he2024id} aim to integrate visual embeddings into diffusion models at training time in a data-driven manner, enabling personalization without additional test-time tuning. 
Despite their progress, how to adopt these approaches to simultaneously process multiple concepts in videos to keep both concept fidelity and factorization in a feed-forward approach remains challenging.

\begin{figure}[t]
\centering
\small 
\begin{minipage}[t]{1.0\linewidth}
\centering
\includegraphics[width=0.9\columnwidth]{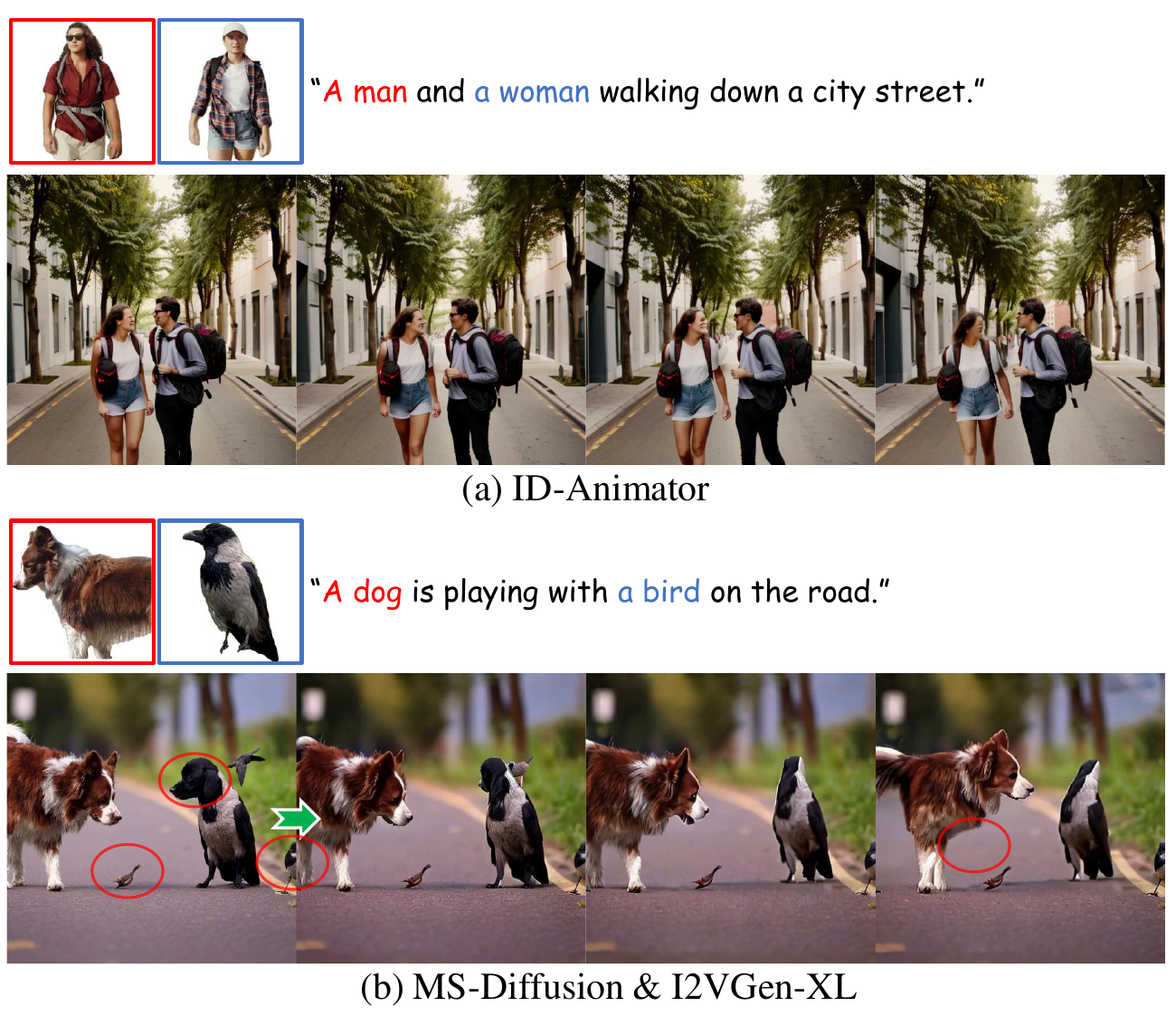}
\end{minipage}
\centering
\vspace{-0.7cm}
\caption{Directly applying single-concept method cannot handle the MCVC task, while the naive solution by combining multi-concept image generation and image-to-video generation models can also hardly create satisfactory customized results.}
\vspace{-0.5cm}
\label{fig:motivation} 
\end{figure}

% 3. 本文提出的ConceptMaster旨在探索MCVC这个任务，有两大难点
% 第一个难点，现在的单ID方法处理不了多IDs任务，会mixing
In this paper, we study the unsolved challenging problem of Multi-Concept Video Customization (MCVC) without test-time tuning, which introduces two critical difficulties: 
1) The identity decoupling problem, unlike single-concept processing, the MCVC task demands not only representing every concept individually based on the given multiple references, but also precisely differentiating the attributes across them in generated videos.
% 直接用现有方法解决不了MCVC
Simply adopting existing pretrain-based approaches often leads to the conflation of visual concepts, inadvertently blending attributes from distinct individuals. This issue becomes more pronounced when dealing with concepts that contain similar attributes.
% 而多id图片+I2V的方式也不可以
A naive composite method is to firstly apply multi-concept image customization based on multiple references, and then input the generated image into image-to-video (I2V) models~\cite{zhang2023i2vgen} for animation. 
However, this approach will be simultaneously subject to the representation and decoupling capability of two models, easily resulting in inferior generation quality and concept consistency.
% motivation figure 重要
As illustrated in Fig.~\ref{fig:motivation}, both these two solutions can neither represent each concept well, nor clearly decouple them across visual appearances, resulting in unacceptable customized videos.
% 第二个难点，缺乏MCVC数据
2) The scarcity of suitable high-quality MCVC datasets. Ideally, training such a customization model requires extensive videos featuring diverse concepts, accompanied by precise textual descriptions and reference images of each entity. 
Current data sources significantly fall short of these requirements, while how to collect large-scale, pairwise video-entity data remains extremely challenging due to the 
accurate extraction of multiple concepts contained in videos across vast diversity of both visual and textual concepts.

% motivation figure

% 4. 我们提出ConceptMaster (learning decoupled multi-concept embeddings and injecting in a standalone manner)
% 提出了ConceptMaster
To overcome these challenges, we propose \textbf{ConceptMaster} (see Fig.~\ref{fig:teaser}), an effective MCVC method that can effectively maintain the fidelity of multiple concepts and address the identity decoupling problem, even for highly similar concepts.
Unlike previous approaches to integrate visual embeddings with textual counterparts~\cite{li2024blip, li2024photomaker, jiang2024videobooth}, or directly aggregate visual embeddings as a whole into diffusion models~\cite{ye2023ip, wang2024ms, he2024id}, our key insight is to learn the decoupled multi-concept embeddings and inject into diffusion transformer models in a standalone manner.
Specifically, the process includes: 1) Extracting comprehensive visual embeddings from given reference images, where we initially extract dense visual tokens by the CLIP image encoder~\cite{radford2021learning}, and integrate a learnable query transformer (Q-Former) network~\cite{li2023blip} to better represent comprehensive visual embeddings and align with the diffusion model space. 
2) Incorporating visual representation with corresponding text description of every concept, where we propose the Decouple Attention Module (DAM) to conduct intra-pair attention to separately bind the extracted visual embedding with corresponding textual embedding for each concept, the process could effectively capture semantic differences across multiple concepts while maintain concept-specific uniqueness. 
3) Introducing a novel multi-concept embeddings injection strategy, where we firstly composite the multi-concept embeddings and then inject them using an individual Multi-Concept Injector (MC-Injector), which is a standalone cross-attention layer, into the diffusion transformer models without affecting the original textual cross-attention. This strategy separates the functionality of the original textual cross-attention from the learning process of the newly injected composite multi-concept embeddings, effectively enhancing the representation of multiple identities.
The designed ConceptMaster could efficiently create high-fidelity customized videos during inference without additional parameter tuning, which significantly provides the potential for the practicality of real-world applications.

% 5. MC-Bench提出，以及多个评测维度
% 我们构建的数据管线打了1.3m+数据
Furthermore, to address the scarcity of suitable MCVC data, we carefully establish a data collection pipeline, which could collect high-quality MCVC data that precisely extract entity images and corresponding text descriptions of diverse concepts in videos.
By utilizing this pipeline, we collect over 1.3 million video-entity pairs spanning diverse conceptual domains, including humans, livings, and various object categories.
% MC-Bench提出
To further facilitate the evaluation, we introduce a multi-concept evaluation set to comprehensively validate this task from 1) concept fidelity, 2) effectiveness of identity decoupling, and 3) video generation quality across six distinct multi-concept composition scenes.
% 6. contributions (as follows)
Overall, our key contributions can be summarized:
\begin{itemize}
    \item We propose ConceptMaster, a novel multi-concept video customization framework to personalize video generation based on user-defined concepts. It effectively addresses the identity decoupling problem while ensuring every concept fidelity, even for highly similar concepts.
    
    \item We present a novel strategy of learning decoupled multi-concept embeddings and injecting them into diffusion models without influencing the original attention operations, which effectively guarantees the fidelity of various concepts in customized videos.
    
    \item We introduce a dedicated data construction pipeline that enables the collection of high-quality multi-concept video-entity pairs across diverse concepts, which effectively address the scarcity of high-quality MCVC data. 
    
    \item We collect a multi-concept evaluation set that could comprehensively validate video customization performance from six distinct concept composition scenarios and various dimensions including concept fidelity, identity decoupling and video quality. Extensive experiments demonstrate the superiority of ConceptMaster in video customization. 
\end{itemize}

\section{Related Work}

%%%%%%%%%%%%%%%%%%%%%%%%%%%%%%%%%%%%%%%%%%%%%%%
% 1.
% 
\subsection{Foundation Text-to-Video Diffusion Models}
The rapid development of text-to-video (T2V) models has been phenomenal. Early works in T2V diffusion models such as AnimateDiff~\cite{guo2023animatediff}, VideoCrafter~\cite{chen2024videocrafter2} and ModelScope~\cite{wang2023modelscope} are mainly based on latent diffusion models~\cite{rombach2022high} with UNet backbones~\cite{ronneberger2015u}. 
By using transformers~\cite{vaswani2017attention} as the backbone of diffusion models, such as Diffusion Transformers (DiT)~\cite{peebles2023scalable}, SORA~\cite{videoworldsimulators2024}, and other transformer-based variants~\cite{hong2022cogvideo, pku_yuan_lab_and_tuzhan_ai_etc_2024_10948109, opensora}, advanced T2V models have scaled parameters and demonstrated impressive capabilities in generating realistic, long-range, and physically consistent videos. This advancement significantly expands the possibilities for content generation.

%%%%%%%%%%%%%%%%%%%%%%%%%%%%%%%%%%%%%%%%%%%%%%%
% 2.
\subsection{Image-based Concept Customization}
Customization in diffusion models enables users to provide reference images to generate results retain the given identities. These customization methods are primarily categorized into tuning-based and pretrain-based approaches. 
Early represented tuning-based methods~\cite{gal2022image, ruiz2023dreambooth} are designed to online-optimize word embeddings or weights of diffusion models when new reference images are provided by users, which are constrained by consuming time and manually collecting training samples. 
Pretrain-based methods~\cite{wei2023elite, gal2023encoder, li2024blip, chen2024anydoor, ye2023ip, xiao2024fastcomposer, li2024photomaker, wang2024instantid} usually train an encoder on certain concept datasets to learn visual representation for conditional diffusion generation process. Some works primarily focus on general-domain concept customization~\cite{wei2023elite, gal2023encoder, li2024blip, chen2024anydoor, ye2023ip}, while others mainly aim at human face identity scenarios~\cite{xiao2024fastcomposer, li2024photomaker, wang2024instantid}.
While aforementioned methods mainly customized single provided concept, the problem also extends to process multiple references. For example, CustomDiffusion~\cite{kumari2023multi} optimizes additional multiple key-value pairs in cross-attention. SSR-Encoder~\cite{zhang2024ssr} aligns query inputs with image patches and preserves fine features of the subjects. MS-Diffusion~\cite{wang2024ms} pretrains a grounding resampler and generates images with bounding box layout guidance. 
These approaches remarkably promote the development of image customization.

%%%%%%%%%%%%%%%%%%%%%%%%%%%%%%%%%%%%%%%%%%%%%%%
% 3.
\subsection{Video-based Concept Customization}
Pretrained-based multi-concept customized video generation raises little attention. Preliminary methods~\cite{wei2024dreamvideo, jiang2024videobooth, he2024id, huang2024story3d} predominantly focus on single-concept scenarios.
DreamVideo~\cite{wei2024dreamvideo} employs a tuning-based approach to simultaneously customize identities and motion. 
Videobooth~\cite{jiang2024videobooth} simply utilizes Grounded-SAM~\cite{kirillov2023segment, liu2023grounding, ren2024grounded} to extract foreground information and tags from the first frame of each video from WebVid dataset~\cite{bain2021frozen} including nine categories as training data, and further trains a coarse-to-fine visual embedding on that data. 
In contrast, ID-Animator~\cite{he2024id} leverages the CelebV dataset~\cite{zhu2022celebv} to construct a face identity dataset, and integrates pre-trained IP-Adapter~\cite{ye2023ip} with AnimateDiff~\cite{guo2023animatediff} for joint optimization. 
However, neither the data collection methods nor the proposed models targeting for single-concept customization are sufficient to directly transfer into multi-concept scenarios. 
ConceptMaster, on the other hand, could solve the challenging MCVC task well in a feed-forward manner, we believe that ConceptMaster has substantially promoted the development of video customization and paved the way for its future.

\section{Preliminary: Diffusion Transformer Models for Text-to-Video Generation}
\label{method:Preliminary}

%%%%%%%%%%%%%%%%%%%%%%%%%%%%%%%%%%%%%%%%%%%%%%%
% 1.
Transformer-based text-to-video diffusion models demonstrate huge potential on video content generation. Our ConceptMaster is built upon a transformer-based latent diffusion model, which employs a 3D Variational Autoencoder (VAE)~\cite{kingma2013auto} to transform videos from the pixel level to a latent space. Each basic transformer block consists of 2D spatial self-attention, 3D spatial-temporal self-attention, text cross-attention, and feed-forward network (FFN). The text prompt embedding $c_{text}$ for cross-attention is obtained by T5 encoder $\mathcal{E}_{T5}$~\cite{raffel2020exploring}. We use Rectified Flow~\cite{liu2022flow, esser2024scaling} to define a probability flow ordinary differential equation (ODE), which transfers the clean data  $z_0$ to a noised data $z_t$ with straight path $z_t = (1-t)z_0 + t\epsilon$ at timestep $t$, where $\epsilon$ is a normal gaussian noise. The diffusion transformer output directly parameterizes the $v_{\Theta}(z_t, t, c_{text})$ to regress velocity $(z_1 - z_0)$ with the Flow Matching objective~\cite{lipman2022flow}:
\begin{equation}\label{eq:loss}
    \mathcal{L}_{LCM}=\mathbb{E}_{t,z_0,\epsilon} ||v_{\Theta}(z_t,t,c_{text})-(z_1-z_0)||_2^2.
\end{equation}

\section{ConceptMaster}

\begin{figure*}[t]
	\centering
	\includegraphics[width=0.78\textwidth]{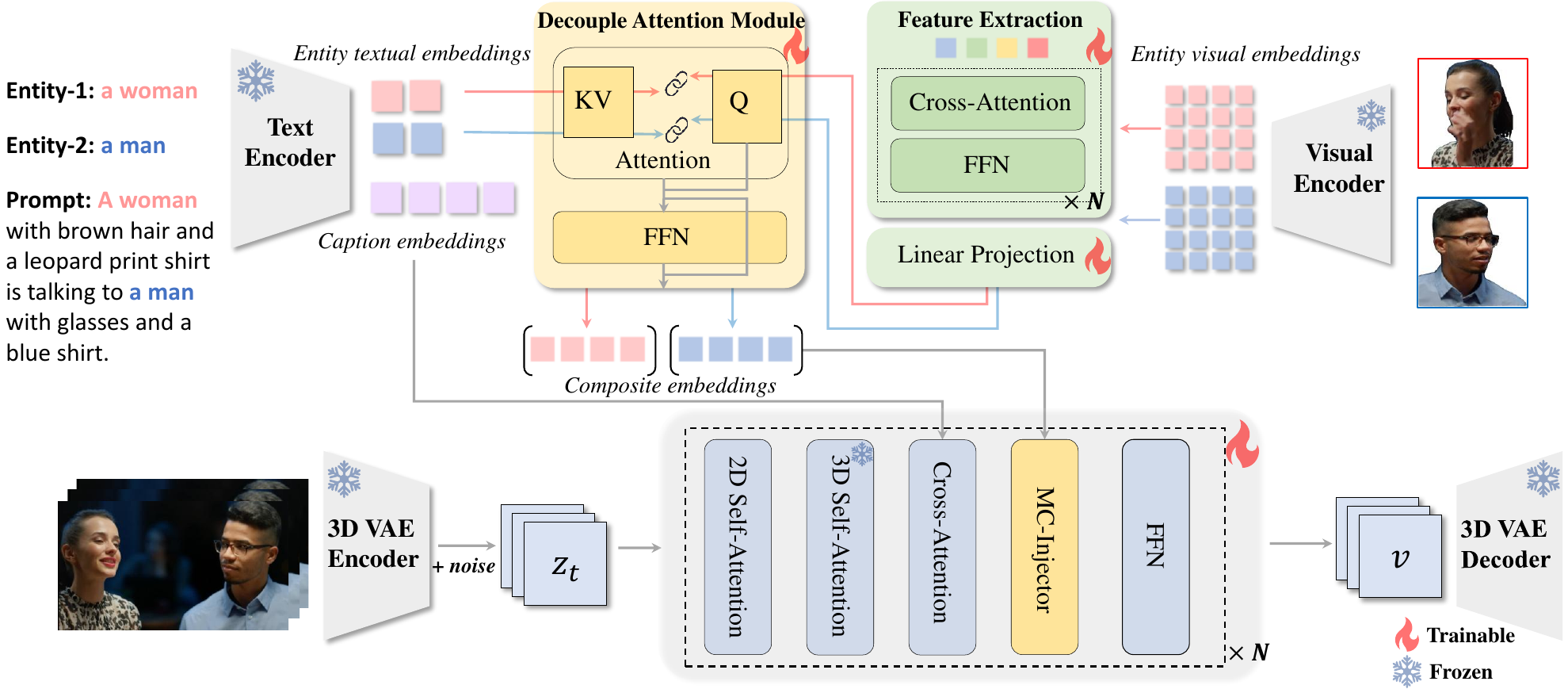}
        \vspace{-2mm}
	\caption{
    Overview of ConceptMaster framework. 
    Given a caption along with a set of concept images and their semantic labels, we firstly extract comprehensive visual concept representations with the CLIP image encoder and a learnable Q-Former, then bind the visual representations with corresponding text embeddings of each concept through the Decouple Attention Module (DAM). Finally, the multi-concept visual-text embeddings are injected into the diffusion transformer models with the Multi-Concept Injector (MC-Injector).
    }
	\label{fig:framework}
	\vspace{-5mm}
\end{figure*}

%%%%%%%%%%%%%%%%%%%%%%%%%%%%%%%%%%%%%%%%%%%%%%%
% 2.
\subsection{Multi-Concept Video Customization}

\textbf{Problem:} Given a caption $T$ describing a video, along with a set of concept images $ \{ X_i | i = 1 \ldots N\}$ and their corresponding labels $\{Y_i | i = 1 \ldots N\}$ (\eg, \emph{a man} and \emph{a woman} with their respective images), where $N$ represents the number of distinct concepts, the task of Multi-Concept Video Customization (MCVC) aims to generate high-quality videos that incorporate all image-defined visual concepts while aligning them with the given descriptive caption $T$. Each concept should maintain its identity as the provided images while precisely expressing its semantic behavior as described in the caption.
We define the paire images and label for each concept as the \textit{intra-pair} customized concept for convenience of expression.
\\
% 收集数据
\textbf{Overview:} To achieve this goal, we first meticulously design a data collection pipeline, resulting in the creation of a dataset comprising over $1.3$ million high-quality MCVC samples. These training videos provide precise information about each entity's image and corresponding text description. Additionally, we incorporate several existing single-concept image and video datasets to further enhance the concept representation.
Afterwards, in order to generate videos that could effectively maintain the fidelity of each concept and decouple multiple visual representation, we firstly extract thoughtful visual embeddings of given reference images, and then design the Decouple Attention Module (DAM) to perform intra-pair attention across paired image-label features, achieving multi-modal representation for each identity. 
Subsequently, we combine every multi-modal concept embedding into the composite ones, and further introduce a Multi-Concept Injector (MC-Injector) in a cross-attention manner to embed the multi-modal composite representation into the diffusion transformer models, where the composite features serve as keys and values.
In Fig.~\ref{fig:framework}, we demonstrate the overview framework of our proposed ConceptMaster.

%%%%%%%%%%%%%%%%%%%%%%%%%%%%%%%%%%%%%%%%%%%%%%%
% 3.
\subsection{Decoupling and Injecting Concept Embeddings}
\label{method:MCInjection}

% 3.1.
\noindent \textbf{Visual Concept Representation Extraction.} 
% CLIP提取16×16×768
To enable the model to process multiple concepts with high fidelity, we need to obtain reasonable visual representation from the concept images  $ \{ X_i | i = 1 \ldots N\}$. We opt to use the CLIP image encoder $\mathcal{E}_{img}$~\cite{radford2021learning} to extract the last layer output as dense visual tokens with shapes $16\times16\times768$, \ie, $\{f_i | f_i = \mathcal{E}_{img}(X_i), i = 1 \ldots N\}$. These tokens have demonstrated more complete visual representation of image conditions~\cite{ye2023ip, shi2024instantbooth, wei2023elite}. 
% CLIP提取的不能直接注入，不足以对齐基模型space
However, directly applying these dense visual tokens in diffusion generation often achieves inadequate alignment with representation space of diffusion models, results in unsatisfactory visual fidelity.
% QFormer可以更好对齐基模型context
To prevent such trivial visual conditions injection and achieve better alignment with the diffusion transformer context, we integrate a learnable Q-Former architecture $\mathcal{Q}$, which comprises stacked cross-attention layers and FFN~\cite{li2023blip, li2024blip, xing2025dynamicrafter}. We utilize the dense visual tokens as a key-value corpus and employ the Q-Former to query these tokens $\{x_i | x_i = \mathcal{Q}(f_i), i = 1 \ldots N\}$, thereby extracting comprehensive visual semantic representation.

% 3.2.
\noindent \textbf{Decoupling Intra-Pair Embeddings.}
% 充分的visual representation已经得到
After obtaining the appropriate visual representation, we integrate the corresponding text labels to create visual-text aligned concept representation.
% FastComposer/Photomaker/BLIPDiffusion对文本的操作我们为什么不行
% 就直接说我们希望最大化每个concept的对应标签信息
While previous works~\cite{xiao2024fastcomposer, li2024photomaker} directly combine the visual representation with the corresponding word from the caption embedding $c_{text} = \mathcal{E}_{\text{text}}(T)$, we hope to fully leverage the textual label information associated with related image to enhance the representation specific to each concept.
% 我们如何利用T5
Therefore, unlike these approaches, we employ T5-encoder $\mathcal{E}_{T5}$ to encode each concept label individually to obtain the text representation $\{y_i | y_i = \mathcal{E}_{T5}(Y_i), i = 1 \ldots N\}$. Subsequently, we introduce the Decouple Attention Module (DAM) to fuse each pair of the visual and text label embedding $\{(x_i, y_i) | i = 1 \ldots N\}$. The DAM operation can be formulated as:

% \vspace{-5mm}
% \begin{equation}\label{eq:DAM}
% \left \{ 
% \begin{aligned}
%  & Q_i = W_{Q} \cdot x_i; K_i = W_K \cdot y_i; V_i = W_V \cdot y_i, \\
%  & Attention(Q_i, K_i, V_i) = \text{Softmax}(\frac{Q_iK_i^{T}}{\sqrt{d}}) \cdot V_i , \\
%  & c_i = \text{FFN}(Attention(Q_i, K_i, V_i)) 
% \end{aligned} 
% \right.
% \end{equation}
\vspace{-5mm}
\begin{equation}\label{eq:DAM}
\left \{ 
\begin{aligned}
 & Q_i = W_{Q} \cdot x_i; K_i = W_K \cdot y_i; V_i = W_V \cdot y_i, \\
 & Attention(Q_i, K_i, V_i) = \text{Softmax}(\frac{Q_iK_i^{T}}{\sqrt{d}}) \cdot V_i , \\
 & b_i = Q_i + Attention(Q_i, K_i, V_i) \\
 & c_i = b_i + \text{FFN}(b_i) 
\end{aligned} 
\right.
\end{equation}

where $W_Q$, $W_K$, and $W_V$ are projection matrices, $d$ is the embedding dimension, and FFN is a two-layer multi-layer perceptron (MLP) with \textit{GLUE}~\cite{wang2018glue} as the middle activation function. The residual connection~\cite{he2016deep} is existed in both attention and MLP layers. With the designed DAM, every visual representation could integrate its corresponding textual label to serve as the visual-text aligned representation for the diffusion transformer models.

% 3.3.
\noindent \textbf{Composite Multi-Concept Representation Injection.} 
After obtaining the multi-modal representation of each pair $\{ c_i | i=1 \ldots N\}$, we firstly concatenate all concept embeddings into a composite one, where $D$ is the dimension of concept embedding:

\vspace{-5mm}
\begin{equation}\label{eq:composite}
    c_{IDs}^* = \text{Concat}(c_1, \ldots, c_N), \quad c_{IDs}^* \in \mathbb{R}^{N\times D}
\end{equation}

Additionally, we design a Multi-Concept Injector (MC-Injector) to encode the composite multi-concept embeddings into the diffusion transformer models. 
Specifically, the MC-Injector is an additional specialized cross-attention layer integrated within each transformer block, positioned after the original text cross-attention layer. 
The additional standalone cross-attention layer can effectively learn the concepts without interference of the original text cross-attention. Comparing with merging the composite embeddings into the original text cross-attention layer, our experiments in Sec.~\ref{exp:Injection} indicate that by interleaving the MC-Injector with the original one could achieve both better decoupling ability and visual fidelity on generated videos.
Finally, the specific diffusion process assisted by the composite embeddings $c_{IDs}^*$ can be formulated as:

\vspace{-5mm}
\begin{equation}\label{eq:new_loss}
    \mathcal{L}_{LCM}=\mathbb{E}_{t,z_0,\epsilon} ||v_{\Theta}(z_t,t,c_{text}, c_{IDs}^*)-(z_1-z_0)||_2^2.
\end{equation}

%%%%%%%%%%%%%%%%%%%%%%%%%%%%%%%%%%%%%%%%%%%%%%%
% data pipe image
\begin{figure*}[t]
	\centering
	\includegraphics[width=1.0\textwidth]{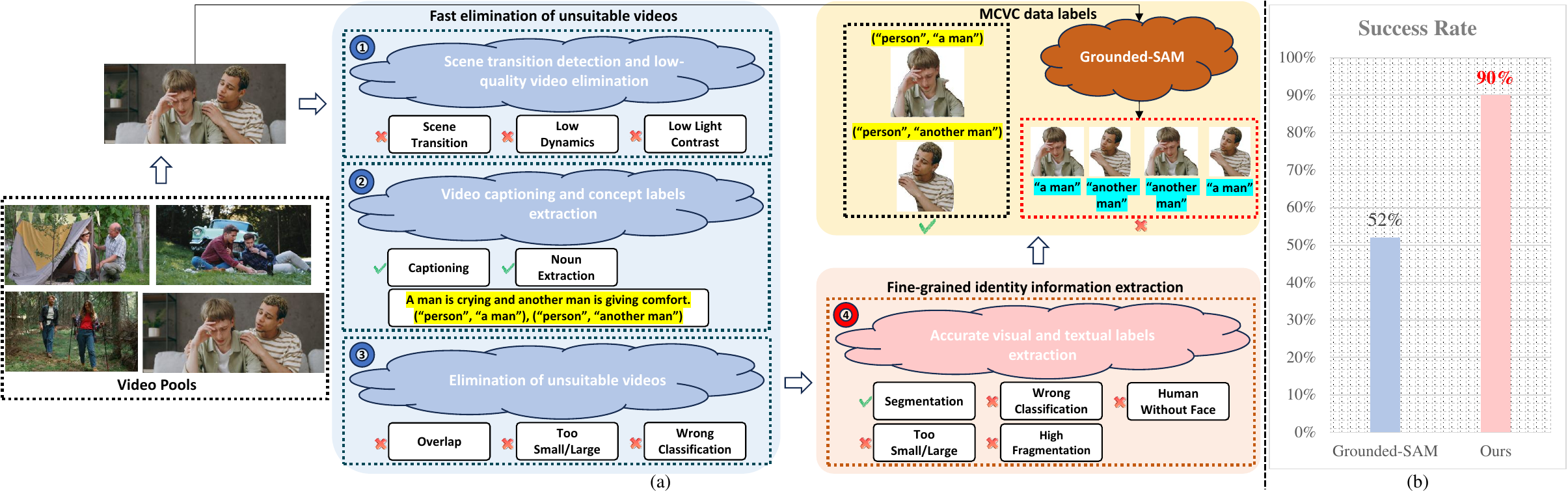}
        % \vspace{-2mm}
        \vspace{-7mm}
	\caption{(a) The overview of multi-concept data collection pipeline. When dealing with complex scenarios that contain concepts with high visual appearance or textual semantic similarity, our data pipeline could still extract precise entity images and corresponding labels, while simply exploit previous methods like Grounded-SAM would introduce a large number of errors and it is difficult to remove these errors through subsequent processing. (b) The success rate of testing videos comparison between Grounded-SAM and our data pipeline.}
	\label{fig:datapipe}
	\vspace{-6mm}
\end{figure*}

% 4.
\subsection{MC-Oriented Video Data Construction}
\label{method:MCVCData}

% 需要高质量MCVC数据
Training a good MCVC model requires high-quality MC-oriented video data.
% 之前方法用Grounded-SAM
Previous studies~\cite{wang2024ms, ma2024subject, jiang2024videobooth} heavily relied on the state-of-the-art open-set object detection methods, such as Grounding-DINO~\cite{liu2023grounding}, to obtain bounding boxes for each concept based on its text label. They then employ the segmentation model SAM~\cite{kirillov2023segment} to extract masks using the bounding boxes as input. 
% Grounded-SAM远远不够
However, the simplistic method is far insufficient for our objectives, as Grounding-DINO, equipped with the CLIP text encoder, often perform poorly in distinguishing similar concepts, especially those that have high visual appearance or textual semantic similarity.
% 质量差的数据不利于MCVC任务
Additionally, incorporating low-quality videos or data not suitable for customization task in training can adversely affect the quality of generated videos.
% 我们的two levels pipeline
Consequently, in order to collect high-quality and large-scale MCVC data, we carefully design the data collection pipeline into two levels: 1) Fast elimination of unsuitable videos, we filtering out low-quality videos that are not unsuitable for the task with time efficiency and low resources. 2) Fine-grained identity information extraction, we guarantee the accuracy of extracted identity reference images and corresponding text labels. We finally collect more than $1.3$ million MCVC data for our ConceptMaster.
Fig.~\ref{fig:datapipe}(a) demonstrates the overview of our dataset collection pipeline.
Additionally, we randomly sample 2000 samples from Panda-2M~\cite{chen2024panda}, and we count the success rate of collect videos. Fig.~\ref{fig:datapipe}(b) demonstrates our designed data pipeline is significantly better than simply using Grounded-SAM. More discussions could be found in appendix.

% 4.1.
\noindent \textbf{Fast elimination of unsuitable videos.} 
\begin{itemize}
    % 1.
    \item \textit{Scene transition detection and low-quality video elimination.} 
    We initially collect more than $6.4$ million videos from Internet as sources. To ensure the basic attributes of our video data are maintained at a high standard, we initially use PySceneDetect~\cite{Castellano_PySceneDetect} to filter out videos that contain scene transitions to maintain the temporal coherence in videos. We also remove videos with low optical flow scores~\cite{teed2020raft} to guarantee the dynamic integrity. Additionally, videos with low light contrast are excluded.

    % 2.
    \item \textit{Video captioning and concept labels extraction.}
    We employ Qwen2-VL~\cite{wang2024qwen2} to produce accurate and concise captions for videos. To extract potential concept entity textual description from the caption, we define a taxonomy of 120 classes, with each class encompassing several sub-words (for instance, the class \textit{dog} includes sub-words such as \textit{dog}, \textit{puppy} and \textit{beagle}). we utilize SpaCy~\cite{Honnibal_spaCy_Industrial-strength_Natural_2020} to extract nouns from the captions, ensuring that these nouns fall within the predefined set of sub-words. The extracted nouns serve as the textual input for text-guided detection and segmentation algorithms. 

    % 3.
    \item \textit{Elimination of unsuitable videos for MCVC task.}
    Since most videos are unsuitable for video customization, we hope to quickly exclude those clearly cannot meet our requirements with minimal resource expenditure and time consuming. 
    For each video, we uniformly sample $10\%$ of the frames and use the extracted nouns to identify entity boxes through text-guided Grounding-DINO. Simultaneously, we apply Non-Maximum Suppression (NMS) to filter out duplicate boxes and remove boxes that are either too large or too small (\eg, areas smaller than $10\%$ or larger than $90\%$ of the video frame size).
    Subsequently, we classify each box using CLIP, eliminating any box if the label classified by CLIP is inconsistent with the original one. If all the boxes are eliminated through this process, the corresponding video will be excluded.
\end{itemize}

% 和加作者一样加下角标
\maketitle
\let\thefootnote\relax\footnotetext{$^1$\href{github repo}{https://github.com/deepinsight/insightface}}

% \renewcommand{\thefootnote}{\arabic{footnote}}
% 4.2.
\noindent \textbf{Fine-grained identity information extraction.} 
\begin{itemize}
    % 1.
    \item \textit{Accurate visual and textual labels extraction.} 
    To accurately extract the region and text label of each identity, we employ the same frame sampling strategy and use LISA~\cite{lai2024lisa}, an MLLM-based~\cite{liu2023visual} segmentor, input by both text prompts and images with strong visual reasoning capabilities, to extract entity masks. LISA provides highly accurate segmentation results, even for similar visual appearance and textual semantics. Those masks are either too large or too small, or with a high degree of fragmentation are removed. We then derive box regions from these masks and remove any misclassified ones through CLIP classification. Additionally, we use FaceAnalysis$^{1}$ to detect all regions belonging to the \textit{person} class, retaining only those that contain face regions (\ie, removing humans where only the body parts are visible).
\end{itemize}

%%%%%%%%%%%%%%%%%%%%%%%%%%%%%%%%%%%%%%%%%%%%%%%
% 5.
\subsection{Joint Training with Auxiliary Datasets}
In addition to the MCVC data we have constructed, we also utilize auxiliary datasets to enhance concept representation. We reproduce the single-concept image dataset from BLIP-Diffusion~\cite{li2024blip} (around 300k) for high-specificity concept enhancement. Furthermore, we incorporate the single-concept video dataset CelebV~\cite{zhu2022celebv} (about 60k) to improve human representation. 
The data sampling ratio of our built data, BLIP-Diffusion and CelebV is 8:1:1.

% \vspace{-1mm}
\section{Experiments}

% comparison image (main)
\begin{figure*}[t]
	\centering
\includegraphics[width=1.0\textwidth]{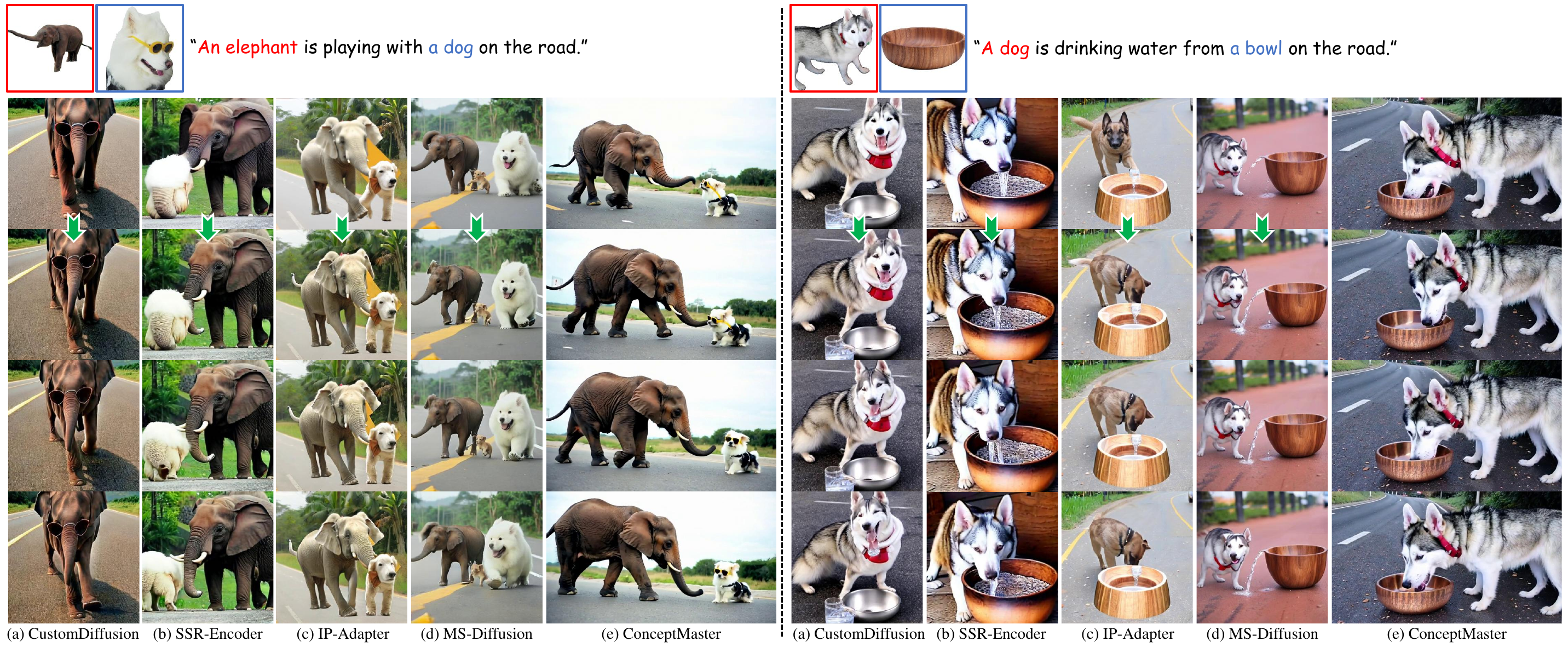}
	\vspace{-7mm}
	\caption{Qualitative comparison on multi-concept customization. When compared to several different methods to conduct the MCVC task, our approach clearly demonstrates superior capabilities on concept fidelity, identity decoupling and caption semantic consistency.}
	\label{fig:comparison}
	\vspace{-1mm}
\end{figure*}

% Tab-1. Comparison table (main)
% 这个是重新填的表
\begin{table*}
\centering
\resizebox{1.0\textwidth}{!}{
\begin{tabular}{c|ccccc|cccc}
\hline
\multirow{2}{*}{Methods} & \multicolumn{5}{c|}{Concept Fidelity and Decoupling Ability} & \multicolumn{4}{c}{Video Quality}  \\ \cline{2-10} 
                 & CLIP-cap$\uparrow$ & CLIP-tag$\uparrow$ & CLIP-I$\uparrow$ & DINO-I$\uparrow$ & CLIP-tag$_{dis}$$\downarrow$ & Motion Smoothness$\uparrow$ & Dynamic Degree$\uparrow$ & Aesthetic Quality$\uparrow$ & Imaging Quality$\uparrow$ \\ \hline

CustomDiffusion  & 26.096 & 21.101 & 0.737 & 0.467 & \underline{16.321} & 0.962 & 15.283 & 0.560 & \underline{0.672} \\
SSR-Encoder  & 23.766 & 21.153 & 0.739 & 0.466 & 16.651 & 0.946 & \underline{21.804} & 0.535 & 0.657 \\
IP-Adapter  & 26.097 & \underline{21.980} & \underline{0.749} & \underline{0.492} & 16.425 & \underline{0.969} & 17.800 & 0.496 & \textbf{0.693} \\
MS-Diffusion  & \underline{28.031} & 21.901 & 0.736 & 0.487 & 16.500 & \textbf{0.978} & 15.849 & \underline{0.571} & 0.624 \\
\hline
ConceptMaster (Ours)  & \textbf{28.246} & \textbf{22.165} & \textbf{0.781} & \textbf{0.584} & \textbf{16.169} & 0.967 & \textbf{26.115} & \textbf{0.572} & 0.657 \\ \hline
\end{tabular}
}
\vspace{-3mm}
\caption{Quantitative comparison with different methods on the introduced multi-concept evaluation set, where \textbf{bold} represents the best result and \underline{underline} represents the second best result.}
\label{table:DifferentMethods}
\vspace{-8mm}
\end{table*}

%%%%%%%%%%%%%%%%%%%%%%%%%%%%%%%%%%%%%%%%%%%%%%%
% 1.
\subsection{Experimental Setup}

\noindent \textbf{Implementation Details.}
The implementation details of ConceptMaster can be found in the supplementary material.

\noindent \textbf{Evaluation Metrics.} 
In order to comprehensively evaluate MCVC methods, we consider  three different dimensions: 
1) Concept fidelity, where we exploit the commonly used CLIP-cap~\cite{radford2021learning} to globally evaluate if the generated videos match the semantics of given video captions. 
2) Decoupling ability, where we utilize LISA~\cite{lai2024lisa} to segment the mask area of each concept in generated videos, and then compute CLIP-I and DINO-I~\cite{caron2021emerging} scores between the original concept images and the mask areas of each concept in generated videos. 
Additionally, we compute the semantic matching scores between each concept label and its corresponding mask area. We refer CLIP-tag and CLIP-tag$_{dis}$ as the similarity and dissimilarity between label and its corresponding area. These four metrics are used to validate the model capacity for decoupling various concepts.
3) Video generation quality, where we adopt the motion smoothness, dynamic degree, aesthetic quality and imaging quality collected as suggested by VBench~\cite{huang2024vbench} to evaluate the generation quality.

%%%%%%%%%%%%%%%%%%%%%%%%%%%%%%%%%%%%%%%%%%%%%%%
% Tab-2. DifferentTech table
\begin{table*}
\centering
\resizebox{1.0\textwidth}{!}{
\begin{tabular}{c|ccccc|cccc}
\hline
\multirow{2}{*}{Methods} & \multicolumn{5}{c|}{Concept Fidelity and Decoupling Ability} & \multicolumn{4}{c}{Video Quality}  \\ \cline{2-10} 
                 & CLIP-cap$\uparrow$ & CLIP-tag$\uparrow$ & CLIP-I$\uparrow$ & DINO-I$\uparrow$ & CLIP-tag$_{dis}$$\downarrow$ & Motion Smoothness$\uparrow$ & Dynamic Degree$\uparrow$ & Aesthetic Quality$\uparrow$ & Imaging Quality$\uparrow$ \\ \hline
                         
Merge Textual and Visual Embeddings  & 27.221 & 21.885 & \underline{0.760} & \underline{0.548} & \underline{16.186} & \textbf{0.969} & 17.025 & \underline{0.538} & \textbf{0.675} \\
IP-Adapter-like  & \underline{28.195} & \underline{22.051} & 0.736 & 0.473 & 16.330 & \textbf{0.969} & \underline{21.957} & 0.530 & 0.652 \\
ConceptMaster (Ours)  & \textbf{28.246} & \textbf{22.165} & \textbf{0.781} & \textbf{0.584} & \textbf{16.169} & \underline{0.967} & \textbf{26.115} & \textbf{0.572} & \underline{0.657} \\ \hline
\end{tabular}
}
\vspace{-3mm}
\caption{Quantitative comparison of different multi-concept embeddings injection manner, \textbf{bold} represents the best result and \underline{underline} represents the second best result. All the methods we compared have been trained on the same data as that used by ConceptMaster.}
\label{table:DifferentTech}
\vspace{-3mm}
\end{table*}

%%%%%%%%%%%%%%%%%%%%%%%%%%%%%%%%%%%%%%%%%%%%%%%
% Tab-3. Ablation table
\begin{table*}
\centering
\resizebox{1.0\textwidth}{!}{
\begin{tabular}{c|ccccc|cccc}
\hline
\multirow{2}{*}{Methods} & \multicolumn{5}{c|}{Concept Fidelity and Decoupling Ability} & \multicolumn{4}{c}{Video Quality}  \\ \cline{2-10} 
                 & CLIP-cap$\uparrow$ & CLIP-tag$\uparrow$ & CLIP-I$\uparrow$ & DINO-I$\uparrow$ & CLIP-tag$_{dis}$$\downarrow$ & Motion Smoothness$\uparrow$ & Dynamic Degree$\uparrow$ & Aesthetic Quality$\uparrow$ & Imaging Quality$\uparrow$ \\ \hline

Without Q-Former  & \textbf{29.312} & 21.876 & 0.726 & 0.416 & 16.379 & 0.963 & 24.984 & 0.518 & \textbf{0.666} \\ 
Without DAM  & \underline{28.916} & 21.981 & 0.730 & 0.439 & 16.365 & \textbf{0.968} & \underline{25.252} & 0.531 & 0.649 \\  
Concate-MLP  & 27.746 & \underline{22.063} & 0.775 & \underline{0.577} & 16.251 & 0.966 & 23.770 & \underline{0.551} & 0.648 \\
Self-Attn  & 28.146 & 22.045 & \underline{0.778} & 0.576 & \underline{16.211} & \textbf{0.968} & 22.864 & 0.550 & 0.655 \\
DAM (Ours)  & 28.246 & \textbf{22.165} & \textbf{0.781} & \textbf{0.584} & \textbf{16.169} & \underline{0.967} & \textbf{26.115} & \textbf{0.572} & \underline{0.657} \\ \hline
\end{tabular}
}
\vspace{-3mm}
\caption{Quantitative comparison of the design choice of Q-Former and DAM modules, \textbf{bold} represents the best result and \underline{underline} represents the second best result. All the methods we compared have been trained on the same data as that used by ConceptMaster.}
\vspace{-5mm}
\label{table:Ablation}
\end{table*}

%%%%%%%%%%%%%%%%%%%%%%%%%%%%%%%%%%%%%%%%%%%%%%%
% 2.

% \subsection{MC-Bench Evaluation}
\subsection{Multi-Concept Evaluation Set}
\label{method:MCBench}

In order to comprehensively evaluate the performance of MCVC methods, we establish a multi-concept evaluation set, including diverse concept composition scenarios of 1) multiple persons, 2) persons with livings, 3) persons with stuffs, 4) multiple livings, 5) livings with stuffs and 6) persons with both livings and stuffs. 
The sample number for each scenario is 40, 40, 40, 30, 30, and 30 respectively (Total 210).
It should be noticed that we manually collect reference images and provide suitable captions for these scenarios, since we hope to eliminate information leakage for evaluation when extracting concepts from videos via the same MCVC data collection pipeline mentioned in Section~\ref{method:MCVCData} (\eg, the caption is \textit{A man and a woman are talking to each other}, and the extracted \textit{man} and \textit{woman} are doing the same thing). All the quantitative experiments are evaluated by this introduced multi-concept evaluation set.
More details are provided in supplementary material.

%%%%%%%%%%%%%%%%%%%%%%%%%%%%%%%%%%%%%%%%%%%%%%%
% 3.
\subsection{Comparing with other methods}

% 比对包括 feed-forward + tuning-based
We compare several open-sourced multi-concept image customization methods~\cite{kumari2023multi, zhang2024ssr, ye2023ip, wang2024ms}, combining with the image-to-video (I2V) generation model I2VGen-XL~\cite{zhang2023i2vgen}, as a naive solution for the MCVC task with our ConceptMaster. 
According to the qualitative results in Fig.~\ref{fig:comparison}, we can see that our ConceptMaster has clear advantages on customizing multiple concepts in videos. 
The naive solution will be subject to the decoupling and representation ability of both two models, the instruction-following capability of I2V models could further influence the quality of generated videos.
In contrast, the end-to-end video customization models could achieve better results.
The quantitative results in Tab.~\ref{table:DifferentMethods} also demonstrates that our method could not only maintain the representation of multiple concepts, but also generate high-quality text-aligned videos.

%%%%%%%%%%%%%%%%%%%%%%%%%%%%%%%%%%%%%%%%%%%%%%%
% DifferentTech image
\begin{figure}[t]
\centering
\small 
\begin{minipage}[t]{1.0\linewidth}
\centering
\includegraphics[width=0.98\columnwidth]{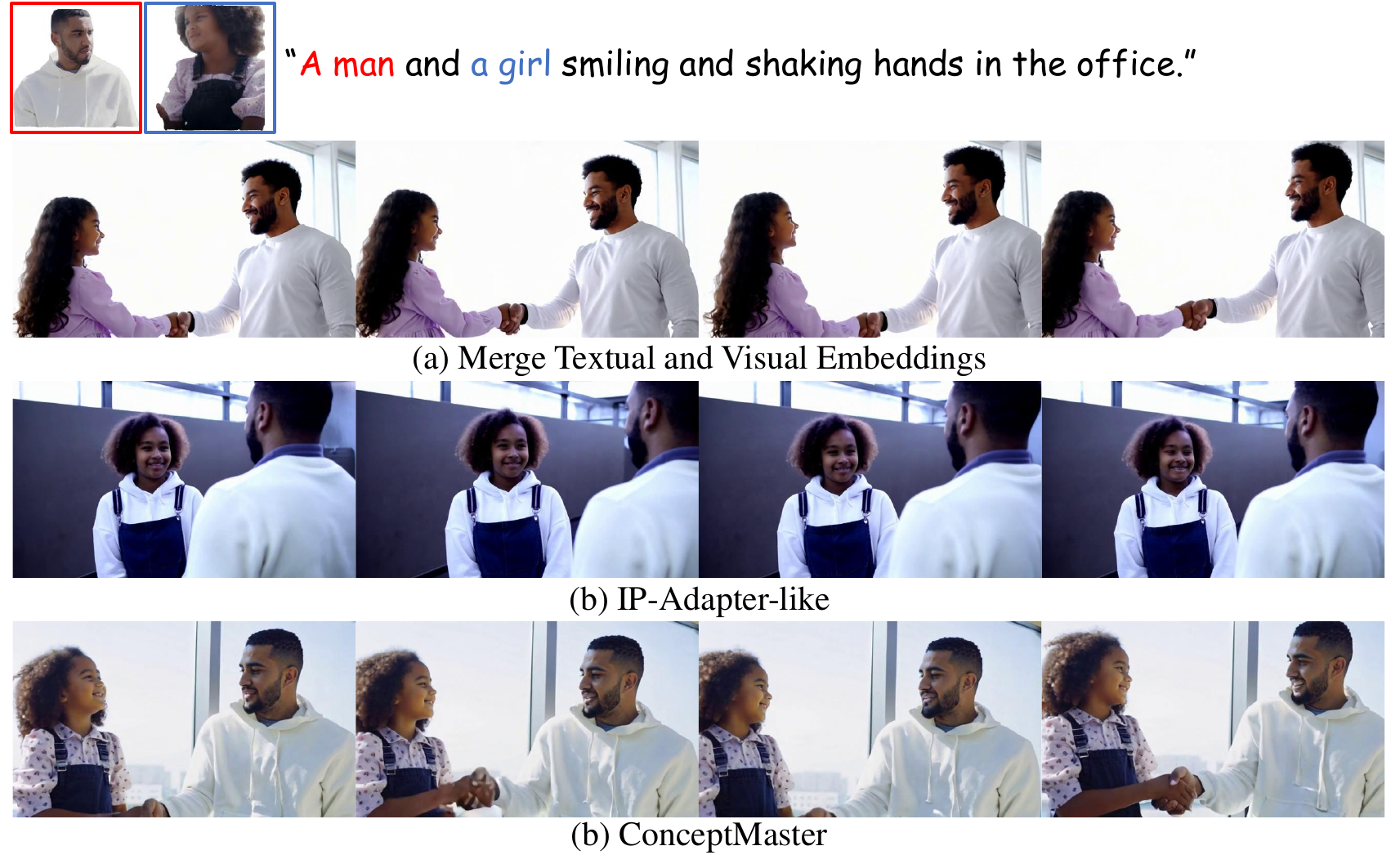}
\end{minipage}
\centering
\vspace{-0.7cm}
\caption{Different injection methods of multi-concept references.}
\vspace{-0.6cm}
\label{fig:Technique} 
\end{figure}

% 4.
\subsection{Multi-Concept Embeddings Injection Manner}
\label{exp:Injection}

% 设计理念
In section~\ref{method:MCInjection}, our ConceptMaster introduces a standalone MC-Injector to integrate the multi-concept visual-text aligned representation into the diffusion models. Some previous methods, represented as BLIP-Diffusion~\cite{li2024blip} and IP-Adapter~\cite{ye2023ip}, the former combines visual embeddings with textual caption embeddings as the whole condition representation, while the latter encodes the whole image as visual embeddings and aggregates into models by a decoupled cross-attention layer.
However, in multi-concept scenarios, merging multi-modal features as the whole conditions can make it challenging to distinguish the semantic meanings among different identities.
In addition, integrating all visual concepts on one image is not the optimal choice for multi-concept representation, especially when they contain similar visual appearances. 
We conduct the above two integration approaches of the multi-concept embeddings on our text-to-video generation models with the same training data for ConceptMaster.
In Tab.~\ref{table:DifferentTech} and Fig.~\ref{fig:Technique}, we can see that when customizing multiple concepts, both these two methods can hardly maintain the concept fidelity and deal with the identity decoupling problem. 
Additionally, when merging the textual and visual embeddings, the dynamic degree is significantly reduced, as the original text cross-attention layer is influenced by the additional visual embeddings.
Therefore, our designed ConceptMaster adopts the optimal solution to inject the decoupled multi-concept embeddings into the diffusion models.

%%%%%%%%%%%%%%%%%%%%%%%%%%%%%%%%%%%%%%%%%%%%%%%
% Ablation image
\begin{figure}[t]
\centering
\small 
\begin{minipage}[t]{1.0\linewidth}
\centering
\includegraphics[width=0.98\columnwidth]{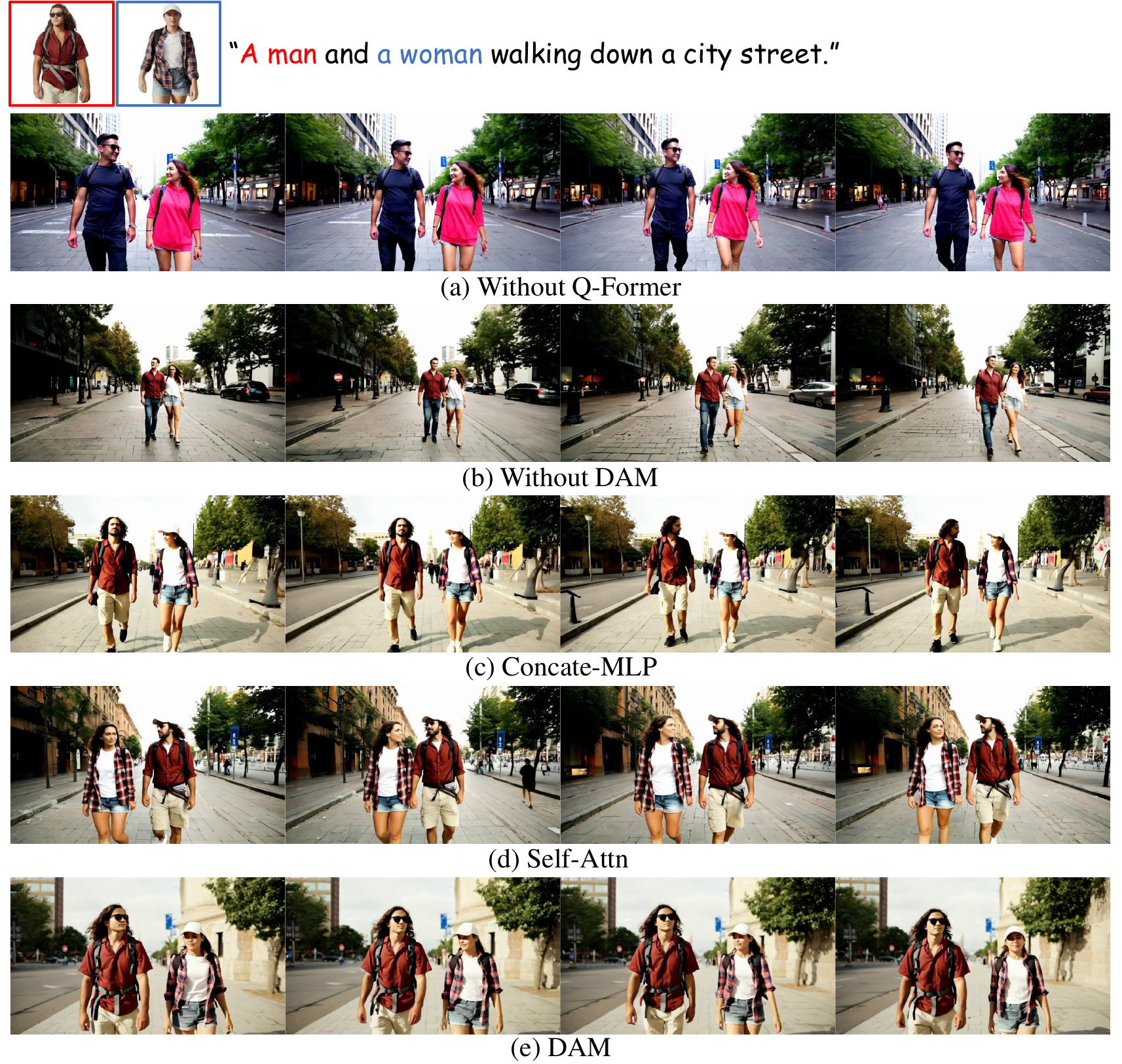}
\end{minipage}
\centering
\vspace{-0.7cm}
\caption{Demonstration of the effectiveness of the Q-Former and DAM modules.}
\vspace{-0.8cm}
\label{fig:ablation} 
\end{figure}

% 5.
\subsection{Ablation Study}

% ablation of DAM and visual embeddings extraction
% Q-Former design
In section~\ref{method:MCInjection}, our ConceptMaster proposes to firstly utilizes a Q-Former network to integrate the dense visual tokens extracted by CLIP image encoder into the comprehensive visual embeddings. While after simply replacing the Q-Former by an MLP layer, the generated videos cannot capture the appearances of given images, as in Fig.~\ref{fig:ablation}, and the quantitative metrics also largely drop in Tab.~\ref{table:Ablation}.

% DAM design
In addition, in order to demonstrate the effectiveness of the designed DAM module, which is the intra-pair attention module based on paired visual embeddings and textual descriptions representation, we conduct several variants include: 
1) Replacing the Q-Former, where we only use an MLP layer instead.
2) Removing the DAM module, where only the extracted visual embeddings are further injected into the diffusion models, and the textual descriptions are unused. 
3) Replacing the intra-pair cross-attention by firstly concatenating the visual and textual embeddings along channel dimension (double the channel dimension), and integrating to the original channel dimension by an MLP layer. 
4) Conducting the intra-pair self-attention instead of cross-attention, the features conduct self-attention are obtained by directly adding the visual and textual embeddings. 
In Tab.~\ref{table:Ablation} and Fig.~\ref{fig:ablation}, we can see that the proposed DAM module is the optimal design. Cooperating the textual descriptions is significant, which not only enhances the uniqueness of each concept representation, but also assists the alignment of the multi-concept embeddings and the original diffusion model space. 
Furthermore, fusing visual and textual embeddings by the MLP layer is less comprehensive as it does not involve sufficient token-level interaction between them, lowering the concept representation and instruction-following in generated results. 
Additionally, the cross-attention operation is better than self-attention, which maintains consistent visual appearances in videos, while more artifacts would be created by the self-attention operation.

\vspace{-2mm}
\section{Conclusion}
\label{sec:conclusion}
\vspace{-2mm}

In this paper, we introduce ConceptMaster, an innovative framework that effectively addresses the critical issues of identity decoupling while maintaining concept fidelity when customizing multiple identities in videos. 
ConceptMaster introduces a novel strategy for learning decoupled multi-concept embeddings and injecting them into diffusion models in a standalone manner. This strategy ensures the quality of customized videos with multiple identities, even for highly similar visual concepts.
To further address the scarcity of high-quality multi-concept video-entity data, we have established a meticulous data construction pipeline. This pipeline enables the systematic collection of precise multi-concept video-entity data across diverse concepts.
Additionally, we have designed a comprehensive testing set to validate the effectiveness of our model from three critical dimensions across six different concept composition scenarios. 
Extensive experiments demonstrate that ConceptMaster significantly outperforms previous approaches, paving the way for generating personalized and semantically accurate videos across multiple concepts.

% \clearpage
{
    \small
    \bibliographystyle{ieeenat_fullname}
    \bibliography{main}

\begin{thebibliography}{71}
\providecommand{\natexlab}[1]{#1}
\providecommand{\url}[1]{\texttt{#1}}
\expandafter\ifx\csname urlstyle\endcsname\relax
  \providecommand{\doi}[1]{doi: #1}\else
  \providecommand{\doi}{doi: \begingroup \urlstyle{rm}\Url}\fi

\bibitem[Bain et~al.(2021)Bain, Nagrani, Varol, and Zisserman]{bain2021frozen}
Max Bain, Arsha Nagrani, G{\"u}l Varol, and Andrew Zisserman.
\newblock Frozen in time: A joint video and image encoder for end-to-end retrieval.
\newblock In \emph{Proceedings of the IEEE/CVF international conference on computer vision}, pages 1728--1738, 2021.

\bibitem[Blattmann et~al.(2023)Blattmann, Rombach, Ling, Dockhorn, Kim, Fidler, and Kreis]{blattmann2023align}
Andreas Blattmann, Robin Rombach, Huan Ling, Tim Dockhorn, Seung~Wook Kim, Sanja Fidler, and Karsten Kreis.
\newblock Align your latents: High-resolution video synthesis with latent diffusion models.
\newblock In \emph{Proceedings of the IEEE/CVF Conference on Computer Vision and Pattern Recognition}, pages 22563--22575, 2023.

\bibitem[Brooks et~al.(2024)Brooks, Peebles, Holmes, DePue, Guo, Jing, Schnurr, Taylor, Luhman, Luhman, Ng, Wang, and Ramesh]{videoworldsimulators2024}
Tim Brooks, Bill Peebles, Connor Holmes, Will DePue, Yufei Guo, Li Jing, David Schnurr, Joe Taylor, Troy Luhman, Eric Luhman, Clarence Ng, Ricky Wang, and Aditya Ramesh.
\newblock Video generation models as world simulators.
\newblock 2024.

\bibitem[Caron et~al.(2021)Caron, Touvron, Misra, J{\'e}gou, Mairal, Bojanowski, and Joulin]{caron2021emerging}
Mathilde Caron, Hugo Touvron, Ishan Misra, Herv{\'e} J{\'e}gou, Julien Mairal, Piotr Bojanowski, and Armand Joulin.
\newblock Emerging properties in self-supervised vision transformers.
\newblock In \emph{Proceedings of the IEEE/CVF international conference on computer vision}, pages 9650--9660, 2021.

\bibitem[Castellano()]{Castellano_PySceneDetect}
Brandon Castellano.
\newblock {PySceneDetect}.

\bibitem[Chen et~al.(2024{\natexlab{a}})Chen, Zhang, Cun, Xia, Wang, Weng, and Shan]{chen2024videocrafter2}
Haoxin Chen, Yong Zhang, Xiaodong Cun, Menghan Xia, Xintao Wang, Chao Weng, and Ying Shan.
\newblock Videocrafter2: Overcoming data limitations for high-quality video diffusion models.
\newblock In \emph{Proceedings of the IEEE/CVF Conference on Computer Vision and Pattern Recognition}, pages 7310--7320, 2024{\natexlab{a}}.

\bibitem[Chen et~al.(2024{\natexlab{b}})Chen, Siarohin, Menapace, Deyneka, Chao, Jeon, Fang, Lee, Ren, Yang, et~al.]{chen2024panda}
Tsai-Shien Chen, Aliaksandr Siarohin, Willi Menapace, Ekaterina Deyneka, Hsiang-wei Chao, Byung~Eun Jeon, Yuwei Fang, Hsin-Ying Lee, Jian Ren, Ming-Hsuan Yang, et~al.
\newblock Panda-70m: Captioning 70m videos with multiple cross-modality teachers.
\newblock In \emph{Proceedings of the IEEE/CVF Conference on Computer Vision and Pattern Recognition}, pages 13320--13331, 2024{\natexlab{b}}.

\bibitem[Chen et~al.(2024{\natexlab{c}})Chen, Huang, Liu, Shen, Zhao, and Zhao]{chen2024anydoor}
Xi Chen, Lianghua Huang, Yu Liu, Yujun Shen, Deli Zhao, and Hengshuang Zhao.
\newblock Anydoor: Zero-shot object-level image customization.
\newblock In \emph{Proceedings of the IEEE/CVF Conference on Computer Vision and Pattern Recognition}, pages 6593--6602, 2024{\natexlab{c}}.

\bibitem[Choi et~al.(2023)Choi, Choi, Kim, Kim, and Yoon]{choi2023custom}
Jooyoung Choi, Yunjey Choi, Yunji Kim, Junho Kim, and Sungroh Yoon.
\newblock Custom-edit: Text-guided image editing with customized diffusion models.
\newblock \emph{arXiv preprint arXiv:2305.15779}, 2023.

\bibitem[Dehghani et~al.(2024)Dehghani, Mustafa, Djolonga, Heek, Minderer, Caron, Steiner, Puigcerver, Geirhos, Alabdulmohsin, et~al.]{dehghani2024patch}
Mostafa Dehghani, Basil Mustafa, Josip Djolonga, Jonathan Heek, Matthias Minderer, Mathilde Caron, Andreas Steiner, Joan Puigcerver, Robert Geirhos, Ibrahim~M Alabdulmohsin, et~al.
\newblock Patch n’pack: Navit, a vision transformer for any aspect ratio and resolution.
\newblock \emph{Advances in Neural Information Processing Systems}, 36, 2024.

\bibitem[Esser et~al.(2023)Esser, Chiu, Atighehchian, Granskog, and Germanidis]{esser2023structure}
Patrick Esser, Johnathan Chiu, Parmida Atighehchian, Jonathan Granskog, and Anastasis Germanidis.
\newblock Structure and content-guided video synthesis with diffusion models.
\newblock \emph{arXiv preprint arXiv:2302.03011}, 2023.

\bibitem[Esser et~al.(2024)Esser, Kulal, Blattmann, Entezari, M{\"u}ller, Saini, Levi, Lorenz, Sauer, Boesel, et~al.]{esser2024scaling}
Patrick Esser, Sumith Kulal, Andreas Blattmann, Rahim Entezari, Jonas M{\"u}ller, Harry Saini, Yam Levi, Dominik Lorenz, Axel Sauer, Frederic Boesel, et~al.
\newblock Scaling rectified flow transformers for high-resolution image synthesis.
\newblock In \emph{Forty-first International Conference on Machine Learning}, 2024.

\bibitem[Gal et~al.(2022)Gal, Alaluf, Atzmon, Patashnik, Bermano, Chechik, and Cohen-Or]{gal2022image}
Rinon Gal, Yuval Alaluf, Yuval Atzmon, Or Patashnik, Amit~H Bermano, Gal Chechik, and Daniel Cohen-Or.
\newblock An image is worth one word: Personalizing text-to-image generation using textual inversion.
\newblock \emph{arXiv preprint arXiv:2208.01618}, 2022.

\bibitem[Gal et~al.(2023)Gal, Arar, Atzmon, Bermano, Chechik, and Cohen-Or]{gal2023encoder}
Rinon Gal, Moab Arar, Yuval Atzmon, Amit~H Bermano, Gal Chechik, and Daniel Cohen-Or.
\newblock Encoder-based domain tuning for fast personalization of text-to-image models.
\newblock \emph{ACM Transactions on Graphics (TOG)}, 42\penalty0 (4):\penalty0 1--13, 2023.

\bibitem[Guo et~al.(2023)Guo, Yang, Rao, Liang, Wang, Qiao, Agrawala, Lin, and Dai]{guo2023animatediff}
Yuwei Guo, Ceyuan Yang, Anyi Rao, Zhengyang Liang, Yaohui Wang, Yu Qiao, Maneesh Agrawala, Dahua Lin, and Bo Dai.
\newblock Animatediff: Animate your personalized text-to-image diffusion models without specific tuning.
\newblock \emph{arXiv preprint arXiv:2307.04725}, 2023.

\bibitem[Han et~al.(2023)Han, Li, Zhang, Milanfar, Metaxas, and Yang]{han2023svdiff}
Ligong Han, Yinxiao Li, Han Zhang, Peyman Milanfar, Dimitris Metaxas, and Feng Yang.
\newblock Svdiff: Compact parameter space for diffusion fine-tuning.
\newblock \emph{arXiv preprint arXiv:2303.11305}, 2023.

\bibitem[He et~al.(2016)He, Zhang, Ren, and Sun]{he2016deep}
Kaiming He, Xiangyu Zhang, Shaoqing Ren, and Jian Sun.
\newblock Deep residual learning for image recognition.
\newblock In \emph{Proceedings of the IEEE conference on computer vision and pattern recognition}, pages 770--778, 2016.

\bibitem[He et~al.(2024)He, Liu, Qian, Wang, Hu, Cao, Yan, Zhou, and Zhang]{he2024id}
Xuanhua He, Quande Liu, Shengju Qian, Xin Wang, Tao Hu, Ke Cao, Keyu Yan, Man Zhou, and Jie Zhang.
\newblock Id-animator: Zero-shot identity-preserving human video generation.
\newblock \emph{arXiv preprint arXiv:2404.15275}, 2024.

\bibitem[He et~al.(2022)He, Yang, Zhang, Shan, and Chen]{he2022latent}
Yingqing He, Tianyu Yang, Yong Zhang, Ying Shan, and Qifeng Chen.
\newblock Latent video diffusion models for high-fidelity video generation with arbitrary lengths.
\newblock \emph{arXiv preprint arXiv:2211.13221}, 2022.

\bibitem[Ho and Salimans(2022)]{ho2022classifier}
Jonathan Ho and Tim Salimans.
\newblock Classifier-free diffusion guidance.
\newblock \emph{arXiv preprint arXiv:2207.12598}, 2022.

\bibitem[Hong et~al.(2022)Hong, Ding, Zheng, Liu, and Tang]{hong2022cogvideo}
Wenyi Hong, Ming Ding, Wendi Zheng, Xinghan Liu, and Jie Tang.
\newblock Cogvideo: Large-scale pretraining for text-to-video generation via transformers.
\newblock \emph{arXiv preprint arXiv:2205.15868}, 2022.

\bibitem[Honnibal et~al.(2020)Honnibal, Montani, Van~Landeghem, and Boyd]{Honnibal_spaCy_Industrial-strength_Natural_2020}
Matthew Honnibal, Ines Montani, Sofie Van~Landeghem, and Adriane Boyd.
\newblock {spaCy: Industrial-strength Natural Language Processing in Python}.
\newblock 2020.

\bibitem[Hu et~al.(2021)Hu, Shen, Wallis, Allen-Zhu, Li, Wang, Wang, and Chen]{hu2021lora}
Edward~J Hu, Yelong Shen, Phillip Wallis, Zeyuan Allen-Zhu, Yuanzhi Li, Shean Wang, Lu Wang, and Weizhu Chen.
\newblock Lora: Low-rank adaptation of large language models.
\newblock \emph{arXiv preprint arXiv:2106.09685}, 2021.

\bibitem[Huang et~al.(2024{\natexlab{a}})Huang, Qin, Lu, Wang, Huang, Shan, and Zhang]{huang2024story3d}
Yuzhou Huang, Yiran Qin, Shunlin Lu, Xintao Wang, Rui Huang, Ying Shan, and Ruimao Zhang.
\newblock Story3d-agent: Exploring 3d storytelling visualization with large language models.
\newblock \emph{arXiv preprint arXiv:2408.11801}, 2024{\natexlab{a}}.

\bibitem[Huang et~al.(2024{\natexlab{b}})Huang, He, Yu, Zhang, Si, Jiang, Zhang, Wu, Jin, Chanpaisit, et~al.]{huang2024vbench}
Ziqi Huang, Yinan He, Jiashuo Yu, Fan Zhang, Chenyang Si, Yuming Jiang, Yuanhan Zhang, Tianxing Wu, Qingyang Jin, Nattapol Chanpaisit, et~al.
\newblock Vbench: Comprehensive benchmark suite for video generative models.
\newblock In \emph{Proceedings of the IEEE/CVF Conference on Computer Vision and Pattern Recognition}, pages 21807--21818, 2024{\natexlab{b}}.

\bibitem[Jiang et~al.(2024)Jiang, Wu, Yang, Si, Lin, Qiao, Loy, and Liu]{jiang2024videobooth}
Yuming Jiang, Tianxing Wu, Shuai Yang, Chenyang Si, Dahua Lin, Yu Qiao, Chen~Change Loy, and Ziwei Liu.
\newblock Videobooth: Diffusion-based video generation with image prompts.
\newblock In \emph{Proceedings of the IEEE/CVF Conference on Computer Vision and Pattern Recognition}, pages 6689--6700, 2024.

\bibitem[Khachatryan et~al.(2023)Khachatryan, Movsisyan, Tadevosyan, Henschel, Wang, Navasardyan, and Shi]{khachatryan2023text2video}
Levon Khachatryan, Andranik Movsisyan, Vahram Tadevosyan, Roberto Henschel, Zhangyang Wang, Shant Navasardyan, and Humphrey Shi.
\newblock Text2video-zero: Text-toimage diffusion models are zero-shot video generators.
\newblock \emph{arXiv preprint arXiv:2303.13439}, 2023.

\bibitem[Kingma(2013)]{kingma2013auto}
Diederik~P Kingma.
\newblock Auto-encoding variational bayes.
\newblock \emph{arXiv preprint arXiv:1312.6114}, 2013.

\bibitem[Kingma(2014)]{kingma2014adam}
Diederik~P Kingma.
\newblock Adam: A method for stochastic optimization.
\newblock \emph{arXiv preprint arXiv:1412.6980}, 2014.

\bibitem[Kirillov et~al.(2023)Kirillov, Mintun, Ravi, Mao, Rolland, Gustafson, Xiao, Whitehead, Berg, Lo, et~al.]{kirillov2023segment}
Alexander Kirillov, Eric Mintun, Nikhila Ravi, Hanzi Mao, Chloe Rolland, Laura Gustafson, Tete Xiao, Spencer Whitehead, Alexander~C Berg, Wan-Yen Lo, et~al.
\newblock Segment anything.
\newblock In \emph{Proceedings of the IEEE/CVF International Conference on Computer Vision}, pages 4015--4026, 2023.

\bibitem[Kumari et~al.(2023)Kumari, Zhang, Zhang, Shechtman, and Zhu]{kumari2023multi}
Nupur Kumari, Bingliang Zhang, Richard Zhang, Eli Shechtman, and Jun-Yan Zhu.
\newblock Multi-concept customization of text-to-image diffusion.
\newblock In \emph{Proceedings of the IEEE/CVF Conference on Computer Vision and Pattern Recognition}, pages 1931--1941, 2023.

\bibitem[Lab and etc.(2024)]{pku_yuan_lab_and_tuzhan_ai_etc_2024_10948109}
PKU-Yuan Lab and Tuzhan~AI etc.
\newblock Open-sora-plan, 2024.

\bibitem[Lai et~al.(2024)Lai, Tian, Chen, Li, Yuan, Liu, and Jia]{lai2024lisa}
Xin Lai, Zhuotao Tian, Yukang Chen, Yanwei Li, Yuhui Yuan, Shu Liu, and Jiaya Jia.
\newblock Lisa: Reasoning segmentation via large language model.
\newblock In \emph{Proceedings of the IEEE/CVF Conference on Computer Vision and Pattern Recognition}, pages 9579--9589, 2024.

\bibitem[Li et~al.(2024{\natexlab{a}})Li, Li, and Hoi]{li2024blip}
Dongxu Li, Junnan Li, and Steven Hoi.
\newblock Blip-diffusion: Pre-trained subject representation for controllable text-to-image generation and editing.
\newblock \emph{Advances in Neural Information Processing Systems}, 36, 2024{\natexlab{a}}.

\bibitem[Li et~al.(2023)Li, Li, Savarese, and Hoi]{li2023blip}
Junnan Li, Dongxu Li, Silvio Savarese, and Steven Hoi.
\newblock Blip-2: Bootstrapping language-image pre-training with frozen image encoders and large language models.
\newblock In \emph{International conference on machine learning}, pages 19730--19742. PMLR, 2023.

\bibitem[Li et~al.(2024{\natexlab{b}})Li, Cao, Wang, Qi, Cheng, and Shan]{li2024photomaker}
Zhen Li, Mingdeng Cao, Xintao Wang, Zhongang Qi, Ming-Ming Cheng, and Ying Shan.
\newblock Photomaker: Customizing realistic human photos via stacked id embedding.
\newblock In \emph{Proceedings of the IEEE/CVF Conference on Computer Vision and Pattern Recognition}, pages 8640--8650, 2024{\natexlab{b}}.

\bibitem[Lipman et~al.(2022)Lipman, Chen, Ben-Hamu, Nickel, and Le]{lipman2022flow}
Yaron Lipman, Ricky~TQ Chen, Heli Ben-Hamu, Maximilian Nickel, and Matt Le.
\newblock Flow matching for generative modeling.
\newblock \emph{arXiv preprint arXiv:2210.02747}, 2022.

\bibitem[Liu et~al.(2023{\natexlab{a}})Liu, Xia, Zhang, Chen, Xing, Wang, Wang, Yang, and Shan]{liu2023stylecrafter}
Gongye Liu, Menghan Xia, Yong Zhang, Haoxin Chen, Jinbo Xing, Yibo Wang, Xintao Wang, Yujiu Yang, and Ying Shan.
\newblock Stylecrafter: Enhancing stylized text-to-video generation with style adapter.
\newblock \emph{arXiv preprint arXiv:2312.00330}, 2023{\natexlab{a}}.

\bibitem[Liu et~al.(2023{\natexlab{b}})Liu, Li, Wu, and Lee]{liu2023visual}
Haotian Liu, Chunyuan Li, Qingyang Wu, and Yong~Jae Lee.
\newblock Visual instruction tuning.
\newblock \emph{arXiv preprint arXiv:2304.08485}, 2023{\natexlab{b}}.

\bibitem[Liu et~al.(2023{\natexlab{c}})Liu, Zeng, Ren, Li, Zhang, Yang, Jiang, Li, Yang, Su, et~al.]{liu2023grounding}
Shilong Liu, Zhaoyang Zeng, Tianhe Ren, Feng Li, Hao Zhang, Jie Yang, Qing Jiang, Chunyuan Li, Jianwei Yang, Hang Su, et~al.
\newblock Grounding dino: Marrying dino with grounded pre-training for open-set object detection.
\newblock \emph{arXiv preprint arXiv:2303.05499}, 2023{\natexlab{c}}.

\bibitem[Liu et~al.(2022)Liu, Gong, and Liu]{liu2022flow}
Xingchao Liu, Chengyue Gong, and Qiang Liu.
\newblock Flow straight and fast: Learning to generate and transfer data with rectified flow.
\newblock \emph{arXiv preprint arXiv:2209.03003}, 2022.

\bibitem[Luo et~al.(2023)Luo, Chen, Zhang, Huang, Wang, Shen, Zhao, Zhou, and Tan]{luo2023videofusion}
Zhengxiong Luo, Dayou Chen, Yingya Zhang, Yan Huang, Liang Wang, Yujun Shen, Deli Zhao, Jingren Zhou, and Tieniu Tan.
\newblock Videofusion: Decomposed diffusion models for high-quality video generation.
\newblock In \emph{Proceedings of the IEEE/CVF Conference on Computer Vision and Pattern Recognition}, pages 10209--10218, 2023.

\bibitem[Ma et~al.(2024)Ma, Liang, Chen, and Lu]{ma2024subject}
Jian Ma, Junhao Liang, Chen Chen, and Haonan Lu.
\newblock Subject-diffusion: Open domain personalized text-to-image generation without test-time fine-tuning.
\newblock In \emph{ACM SIGGRAPH 2024 Conference Papers}, pages 1--12, 2024.

\bibitem[Peebles and Xie(2023)]{peebles2023scalable}
William Peebles and Saining Xie.
\newblock Scalable diffusion models with transformers.
\newblock In \emph{Proceedings of the IEEE/CVF International Conference on Computer Vision}, pages 4195--4205, 2023.

\bibitem[Radford et~al.(2021)Radford, Kim, Hallacy, Ramesh, Goh, Agarwal, Sastry, Askell, Mishkin, Clark, et~al.]{radford2021learning}
Alec Radford, Jong~Wook Kim, Chris Hallacy, Aditya Ramesh, Gabriel Goh, Sandhini Agarwal, Girish Sastry, Amanda Askell, Pamela Mishkin, Jack Clark, et~al.
\newblock Learning transferable visual models from natural language supervision.
\newblock In \emph{International conference on machine learning}, pages 8748--8763. PMLR, 2021.

\bibitem[Raffel et~al.(2020)Raffel, Shazeer, Roberts, Lee, Narang, Matena, Zhou, Li, and Liu]{raffel2020exploring}
Colin Raffel, Noam Shazeer, Adam Roberts, Katherine Lee, Sharan Narang, Michael Matena, Yanqi Zhou, Wei Li, and Peter~J Liu.
\newblock Exploring the limits of transfer learning with a unified text-to-text transformer.
\newblock \emph{Journal of machine learning research}, 21\penalty0 (140):\penalty0 1--67, 2020.

\bibitem[Ren et~al.(2024)Ren, Liu, Zeng, Lin, Li, Cao, Chen, Huang, Chen, Yan, Zeng, Zhang, Li, Yang, Li, Jiang, and Zhang]{ren2024grounded}
Tianhe Ren, Shilong Liu, Ailing Zeng, Jing Lin, Kunchang Li, He Cao, Jiayu Chen, Xinyu Huang, Yukang Chen, Feng Yan, Zhaoyang Zeng, Hao Zhang, Feng Li, Jie Yang, Hongyang Li, Qing Jiang, and Lei Zhang.
\newblock Grounded sam: Assembling open-world models for diverse visual tasks, 2024.

\bibitem[Rombach et~al.(2022)Rombach, Blattmann, Lorenz, Esser, and Ommer]{rombach2022high}
Robin Rombach, Andreas Blattmann, Dominik Lorenz, Patrick Esser, and Bj{\"o}rn Ommer.
\newblock High-resolution image synthesis with latent diffusion models.
\newblock In \emph{Proceedings of the IEEE/CVF conference on computer vision and pattern recognition}, pages 10684--10695, 2022.

\bibitem[Ronneberger et~al.(2015)Ronneberger, Fischer, and Brox]{ronneberger2015u}
Olaf Ronneberger, Philipp Fischer, and Thomas Brox.
\newblock U-net: Convolutional networks for biomedical image segmentation.
\newblock In \emph{Medical Image Computing and Computer-Assisted Intervention--MICCAI 2015: 18th International Conference, Munich, Germany, October 5-9, 2015, Proceedings, Part III 18}, pages 234--241. Springer, 2015.

\bibitem[Ruiz et~al.(2023)Ruiz, Li, Jampani, Pritch, Rubinstein, and Aberman]{ruiz2023dreambooth}
Nataniel Ruiz, Yuanzhen Li, Varun Jampani, Yael Pritch, Michael Rubinstein, and Kfir Aberman.
\newblock Dreambooth: Fine tuning text-to-image diffusion models for subject-driven generation.
\newblock In \emph{Proceedings of the IEEE/CVF conference on computer vision and pattern recognition}, pages 22500--22510, 2023.

\bibitem[Shi et~al.(2024)Shi, Xiong, Lin, and Jung]{shi2024instantbooth}
Jing Shi, Wei Xiong, Zhe Lin, and Hyun~Joon Jung.
\newblock Instantbooth: Personalized text-to-image generation without test-time finetuning.
\newblock In \emph{Proceedings of the IEEE/CVF Conference on Computer Vision and Pattern Recognition}, pages 8543--8552, 2024.

\bibitem[Singer et~al.(2022)Singer, Polyak, Hayes, Yin, An, Zhang, Hu, Yang, Ashual, Gafni, et~al.]{singer2022make}
Uriel Singer, Adam Polyak, Thomas Hayes, Xi Yin, Jie An, Songyang Zhang, Qiyuan Hu, Harry Yang, Oron Ashual, Oran Gafni, et~al.
\newblock Make-a-video: Text-to-video generation without text-video data.
\newblock \emph{arXiv preprint arXiv:2209.14792}, 2022.

\bibitem[Song et~al.(2020)Song, Meng, and Ermon]{song2020denoising}
Jiaming Song, Chenlin Meng, and Stefano Ermon.
\newblock Denoising diffusion implicit models.
\newblock \emph{arXiv preprint arXiv:2010.02502}, 2020.

\bibitem[Teed and Deng(2020)]{teed2020raft}
Zachary Teed and Jia Deng.
\newblock Raft: Recurrent all-pairs field transforms for optical flow.
\newblock In \emph{Computer Vision--ECCV 2020: 16th European Conference, Glasgow, UK, August 23--28, 2020, Proceedings, Part II 16}, pages 402--419. Springer, 2020.

\bibitem[Vaswani(2017)]{vaswani2017attention}
A Vaswani.
\newblock Attention is all you need.
\newblock \emph{Advances in Neural Information Processing Systems}, 2017.

\bibitem[Villegas et~al.(2022)Villegas, Babaeizadeh, Kindermans, Moraldo, Zhang, Saffar, Castro, Kunze, and Erhan]{villegas2022phenaki}
Ruben Villegas, Mohammad Babaeizadeh, Pieter-Jan Kindermans, Hernan Moraldo, Han Zhang, Mohammad~Taghi Saffar, Santiago Castro, Julius Kunze, and Dumitru Erhan.
\newblock Phenaki: Variable length video generation from open domain textual description.
\newblock \emph{arXiv preprint arXiv:2210.02399}, 2022.

\bibitem[Wang(2018)]{wang2018glue}
Alex Wang.
\newblock Glue: A multi-task benchmark and analysis platform for natural language understanding.
\newblock \emph{arXiv preprint arXiv:1804.07461}, 2018.

\bibitem[Wang et~al.(2023)Wang, Yuan, Chen, Zhang, Wang, and Zhang]{wang2023modelscope}
Jiuniu Wang, Hangjie Yuan, Dayou Chen, Yingya Zhang, Xiang Wang, and Shiwei Zhang.
\newblock Modelscope text-to-video technical report.
\newblock \emph{arXiv preprint arXiv:2308.06571}, 2023.

\bibitem[Wang et~al.(2024{\natexlab{a}})Wang, Bai, Tan, Wang, Fan, Bai, Chen, Liu, Wang, Ge, et~al.]{wang2024qwen2}
Peng Wang, Shuai Bai, Sinan Tan, Shijie Wang, Zhihao Fan, Jinze Bai, Keqin Chen, Xuejing Liu, Jialin Wang, Wenbin Ge, et~al.
\newblock Qwen2-vl: Enhancing vision-language model's perception of the world at any resolution.
\newblock \emph{arXiv preprint arXiv:2409.12191}, 2024{\natexlab{a}}.

\bibitem[Wang et~al.(2024{\natexlab{b}})Wang, Bai, Wang, Qin, Chen, Li, Tang, and Hu]{wang2024instantid}
Qixun Wang, Xu Bai, Haofan Wang, Zekui Qin, Anthony Chen, Huaxia Li, Xu Tang, and Yao Hu.
\newblock Instantid: Zero-shot identity-preserving generation in seconds.
\newblock \emph{arXiv preprint arXiv:2401.07519}, 2024{\natexlab{b}}.

\bibitem[Wang et~al.(2024{\natexlab{c}})Wang, Fu, Huang, He, and Jiang]{wang2024ms}
X Wang, Siming Fu, Qihan Huang, Wanggui He, and Hao Jiang.
\newblock Ms-diffusion: Multi-subject zero-shot image personalization with layout guidance.
\newblock \emph{arXiv preprint arXiv:2406.07209}, 2024{\natexlab{c}}.

\bibitem[Wei et~al.(2023)Wei, Zhang, Ji, Bai, Zhang, and Zuo]{wei2023elite}
Yuxiang Wei, Yabo Zhang, Zhilong Ji, Jinfeng Bai, Lei Zhang, and Wangmeng Zuo.
\newblock Elite: Encoding visual concepts into textual embeddings for customized text-to-image generation.
\newblock In \emph{Proceedings of the IEEE/CVF International Conference on Computer Vision}, pages 15943--15953, 2023.

\bibitem[Wei et~al.(2024)Wei, Zhang, Qing, Yuan, Liu, Liu, Zhang, Zhou, and Shan]{wei2024dreamvideo}
Yujie Wei, Shiwei Zhang, Zhiwu Qing, Hangjie Yuan, Zhiheng Liu, Yu Liu, Yingya Zhang, Jingren Zhou, and Hongming Shan.
\newblock Dreamvideo: Composing your dream videos with customized subject and motion.
\newblock In \emph{Proceedings of the IEEE/CVF Conference on Computer Vision and Pattern Recognition}, pages 6537--6549, 2024.

\bibitem[Wu et~al.(2022)Wu, Ge, Wang, Lei, Gu, Hsu, Shan, Qie, and Shou]{wu2022tune}
Jay~Zhangjie Wu, Yixiao Ge, Xintao Wang, Weixian Lei, Yuchao Gu, Wynne Hsu, Ying Shan, Xiaohu Qie, and Mike~Zheng Shou.
\newblock Tune-a-video: One-shot tuning of image diffusion models for text-to-video generation.
\newblock \emph{arXiv preprint arXiv:2212.11565}, 2022.

\bibitem[Xiao et~al.(2024)Xiao, Yin, Freeman, Durand, and Han]{xiao2024fastcomposer}
Guangxuan Xiao, Tianwei Yin, William~T Freeman, Fr{\'e}do Durand, and Song Han.
\newblock Fastcomposer: Tuning-free multi-subject image generation with localized attention.
\newblock \emph{International Journal of Computer Vision}, pages 1--20, 2024.

\bibitem[Xing et~al.(2025)Xing, Xia, Zhang, Chen, Yu, Liu, Liu, Wang, Shan, and Wong]{xing2025dynamicrafter}
Jinbo Xing, Menghan Xia, Yong Zhang, Haoxin Chen, Wangbo Yu, Hanyuan Liu, Gongye Liu, Xintao Wang, Ying Shan, and Tien-Tsin Wong.
\newblock Dynamicrafter: Animating open-domain images with video diffusion priors.
\newblock In \emph{European Conference on Computer Vision}, pages 399--417. Springer, 2025.

\bibitem[Ye et~al.(2023)Ye, Zhang, Liu, Han, and Yang]{ye2023ip}
Hu Ye, Jun Zhang, Sibo Liu, Xiao Han, and Wei Yang.
\newblock Ip-adapter: Text compatible image prompt adapter for text-to-image diffusion models.
\newblock \emph{arXiv preprint arXiv:2308.06721}, 2023.

\bibitem[Zhang et~al.(2023)Zhang, Wang, Zhang, Zhao, Yuan, Qin, Wang, Zhao, and Zhou]{zhang2023i2vgen}
Shiwei Zhang, Jiayu Wang, Yingya Zhang, Kang Zhao, Hangjie Yuan, Zhiwu Qin, Xiang Wang, Deli Zhao, and Jingren Zhou.
\newblock I2vgen-xl: High-quality image-to-video synthesis via cascaded diffusion models.
\newblock \emph{arXiv preprint arXiv:2311.04145}, 2023.

\bibitem[Zhang et~al.(2024)Zhang, Song, Liu, Wang, Yu, Tang, Li, Tang, Hu, Pan, et~al.]{zhang2024ssr}
Yuxuan Zhang, Yiren Song, Jiaming Liu, Rui Wang, Jinpeng Yu, Hao Tang, Huaxia Li, Xu Tang, Yao Hu, Han Pan, et~al.
\newblock Ssr-encoder: Encoding selective subject representation for subject-driven generation.
\newblock In \emph{Proceedings of the IEEE/CVF Conference on Computer Vision and Pattern Recognition}, pages 8069--8078, 2024.

\bibitem[Zheng et~al.(2024)Zheng, Peng, Yang, Shen, Li, Liu, Zhou, Li, and You]{opensora}
Zangwei Zheng, Xiangyu Peng, Tianji Yang, Chenhui Shen, Shenggui Li, Hongxin Liu, Yukun Zhou, Tianyi Li, and Yang You.
\newblock Open-sora: Democratizing efficient video production for all, 2024.

\bibitem[Zhu et~al.(2022)Zhu, Wu, Zhu, Jiang, Tang, Zhang, Liu, and Loy]{zhu2022celebv}
Hao Zhu, Wayne Wu, Wentao Zhu, Liming Jiang, Siwei Tang, Li Zhang, Ziwei Liu, and Chen~Change Loy.
\newblock Celebv-hq: A large-scale video facial attributes dataset.
\newblock In \emph{European conference on computer vision}, pages 650--667. Springer, 2022.

\end{thebibliography}
}

\clearpage
% 从头开始标
\setcounter{section}{0}
\clearpage
\setcounter{page}{1}
\maketitlesupplementary

\noindent We provide the following contents in supplementary materials:

\begin{enumerate}
    \item Introduction of our text-to-video diffusion transformer models.

    \item Implementation Details of ConceptMaster.

    \item Discussions on Comparison between Our Data Collection Pipeline and Grounded-SAM.

    \item More details of Multi-Concept Evaluation Set.

    \item Comparison Methods Implementation.

    \item More Discussions on Multi-Concept Embeddings Injection.

    \item More Discussions on Ablation Study.

    \item More Qualitative Results Demonstration.
    
\end{enumerate}

% 我们要解释一些正文没有解释清楚的东西
% 我们要和videobooth以及idanimator比较
% 我们要展示更多的比较结果
% 我们要展示更多的设计理念和ablation结果 (设计理念是不是还需要补一张图)
% 我们要展示6类场景的图

%%%%%%%%%%%%%%%%%%%%%%%%%%%%%%%%%%%%%%%%%%%%%%%%%%%%%%%%%%%
% 1.
\section{Introduction of our text-to-video diffusion transformer models}
\begin{figure}[t]
\centering
\small 
\begin{minipage}[t]{1.0\linewidth}
\centering
\includegraphics[width=1\columnwidth]{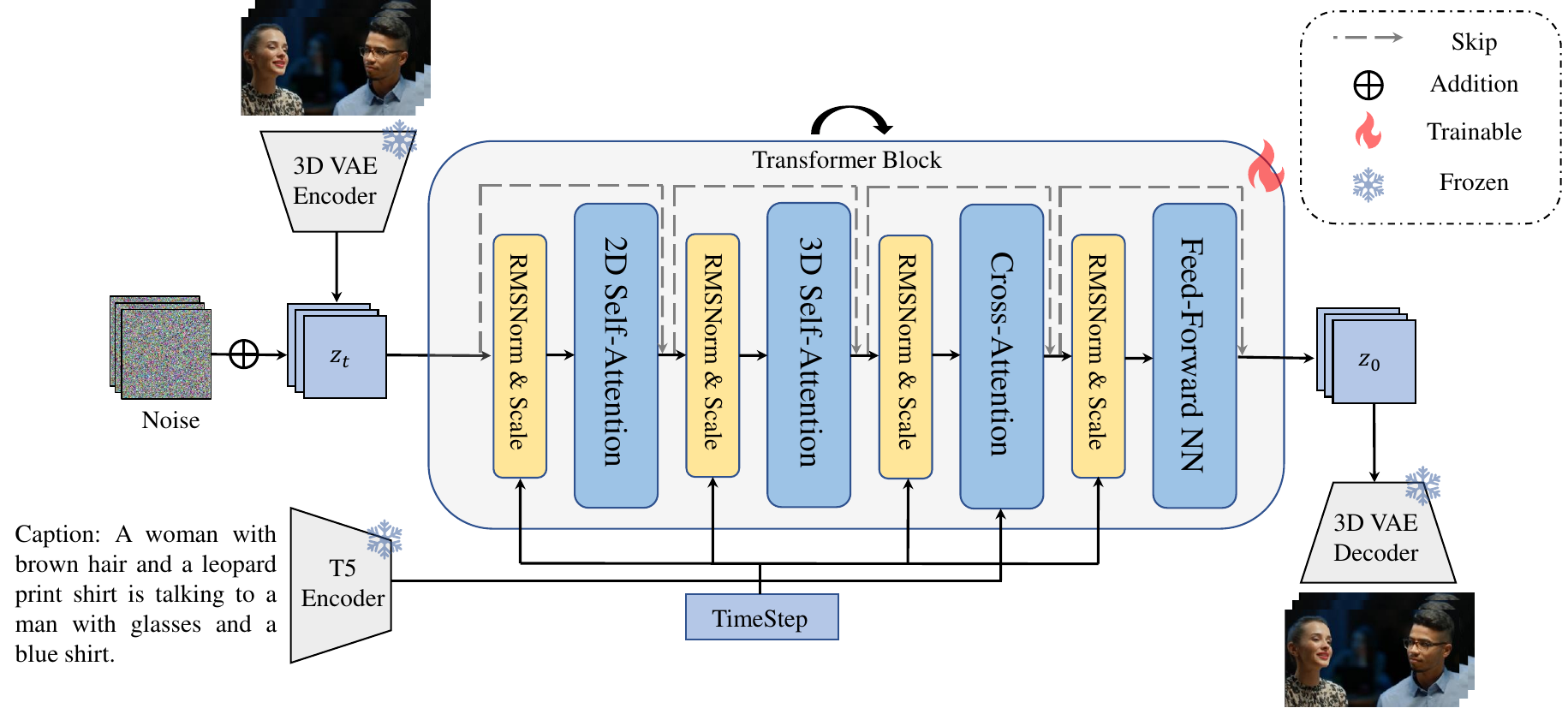}
\end{minipage}
\centering
\caption{Overview framwork of our base text-to-video generation models.}
\label{fig:t2v_model}
\end{figure}

We utilize a transformer-based latent diffusion model as the foundational text-to-video (T2V) generation model, as depicted in Fig.~\ref{fig:t2v_model}. Initially, we employ a 3D Variational Autoencoder (3D-VAE) to transform videos from the pixel space into a latent space, upon which we build a transformer-based video diffusion model. Unlike previous models that rely on UNets or transformers with an additional 1D temporal attention module for video generation, our approach addresses the limitations of spatially-temporally separated designs, which often do not yield optimal results. We replace the 1D temporal attention with 3D self-attention, allowing the model to more effectively perceive and process spatiotemporal tokens. This results in a high-quality and physically coherent video generation model. Specifically, before each attention or feed-forward network (FFN) module, we map the timestep to a scale and apply RMSNorm to the spatiotemporal tokens.

%%%%%%%%%%%%%%%%%%%%%%%%%%%%%%%%%%%%%%%%%%%%%%%%%%%%%%%%%%%
% 2.
\section{Implementation Details of ConceptMaster}

\noindent \textbf{Implementation Details.} 
We train ConceptMaster using our proprietary transformer-based text-to-video diffusion models. 
Initially, we employ the CLIP image encoder~\cite{radford2021learning} as the external vision encoder to extract visual features from reference images.
Subsequently it is followed by the stacked cross-attention and FFN layers, collectively referred to as the Q-Former, and an additional cross-attention layer for the DAM module. 
% 
% The transformer backbone we utilize comprises approximately 1 billion parameters. 
During training, we drop the video captions and reference conditions (both paired images and text descriptions) with probabilities of 50\% and 33\% for classifier-free guidance~\cite{ho2022classifier}, respectively. We freeze the 3D spatiotemporal self-attention layer and fine-tune the other parameters of the transformer to enhance video dynamics. Consequently, the entire transformer backbone, except for the 3D spatiotemporal layer, the Q-Former, and the DAM module, are jointly optimized.
Additionally, inspired by NaViT~\cite{dehghani2024patch}, we adopt the similar strategy of padding the videos to the same height and width with effective attention masks within each batch during training, and the training video segments consist of 77 frames, corresponding to a duration of 5 seconds at 15 frames per second (fps). 
The training process employs the Adam optimizer~\cite{kingma2014adam} and is conducted on 64 NVIDIA H800 GPUs, with a learning rate set to $5\times10^{-6}$ and a global batch size of 256.
During inference, we utilize 100 DDIM steps~\cite{song2020denoising} and set the CFG scale to 7.5, and the inference videos are uniformly resized to a resolution of $384\times672$ pixels.

% \noindent \textbf{Data Processing.} 
% Data augmentation in concept customization is important, which could avoid the copy-paste effect.

% S1.

%%%%%%%%%%%%%%%%%%%%%%%%%%%%%%%%%%%%%%%%%%%%%%%%%%%%%%%%%%%
% 3.
\section{Discussions on Comparison between Our Data Collection Pipeline and Grounded-SAM}
Previous studies typically exploit open-set object detection and segmentation methods, represented by Grounded-SAM~\cite{kirillov2023segment, liu2023grounding, ren2024grounded}, to extract concepts information in source images or videos.
However, we claim that the simplistic method is far insufficient for our objectives, since Grounding-DINO is built on top of the CLIP text encoder, which often perform poorly in distinguishing similar concepts, especially those that have high visual appearance or textual semantic similarity (see Fig~\ref{fig:datapipe}(a)). Additionally, we need to guarantee the quality in a high standard of the source data as well as the extracted concept information (\eg, appropriate size and completeness) to train the model for our task.
We carefully design the data collection pipeline into two levels including: 1) Fast elimination of unsuitable videos and 2) Fine-grained identity information extraction. 
In order to evaluate the effectiveness of our dataset pipeline, we first randomly sample 2000 samples from Panda-2M~\cite{chen2024panda}, and we count the success rate of collect videos. The standard for success rate statistics is based on whether there are errors in the extracted concepts information from a video (\ie, wrong classification label for the concepts). A success is defined as the absence of any errors, while the presence of any error is considered a failure.
We additionally hire 20 experienced workers in data construction to manually evaluate the results of the constructed evaluation samples from Panda-2M. 
We present the success rate statistics in Fig.~\ref{fig:datapipe}(b). The results indicate that the success rate of our designed data pipeline is significantly higher than that achieved by simply using Grounded-SAM. Grounded-SAM exhibits a success rate of only 52\%, which poses challenges in training a robust model capable of representing and decoupling multiple concepts in generated videos, especially when dealing with MCVC data that contains numerous errors. Consequently, our constructed data pipeline is essential for high-quality MCVC data collection. We also hope that our data collection process could provide inspiration for future MCVC works.

%%%%%%%%%%%%%%%%%%%%%%%%%%%%%%%%%%%%%%%%%%%%%%%%%%%%%%%%%%%
% 新增加的 MC-Bench
\section{More details of Multi-Concept Evaluation Set}
As mentioned in section~\ref{method:MCVCData}, we manually collect reference images and give out suitable captions for these scenarios, in order to eliminate information leakage when extracting concepts from videos. We demonstrate the video caption templates for the six different scenarios in Tab.~\ref{table:MCBench}. The \textless ID\textgreater\ will be replaced with concept label.

% 4.
\section{Comparison Methods Implementation}
\begin{figure*}[t]
\centering
\small 
\begin{minipage}[t]{1.0\linewidth}
\centering
\includegraphics[width=1\columnwidth]{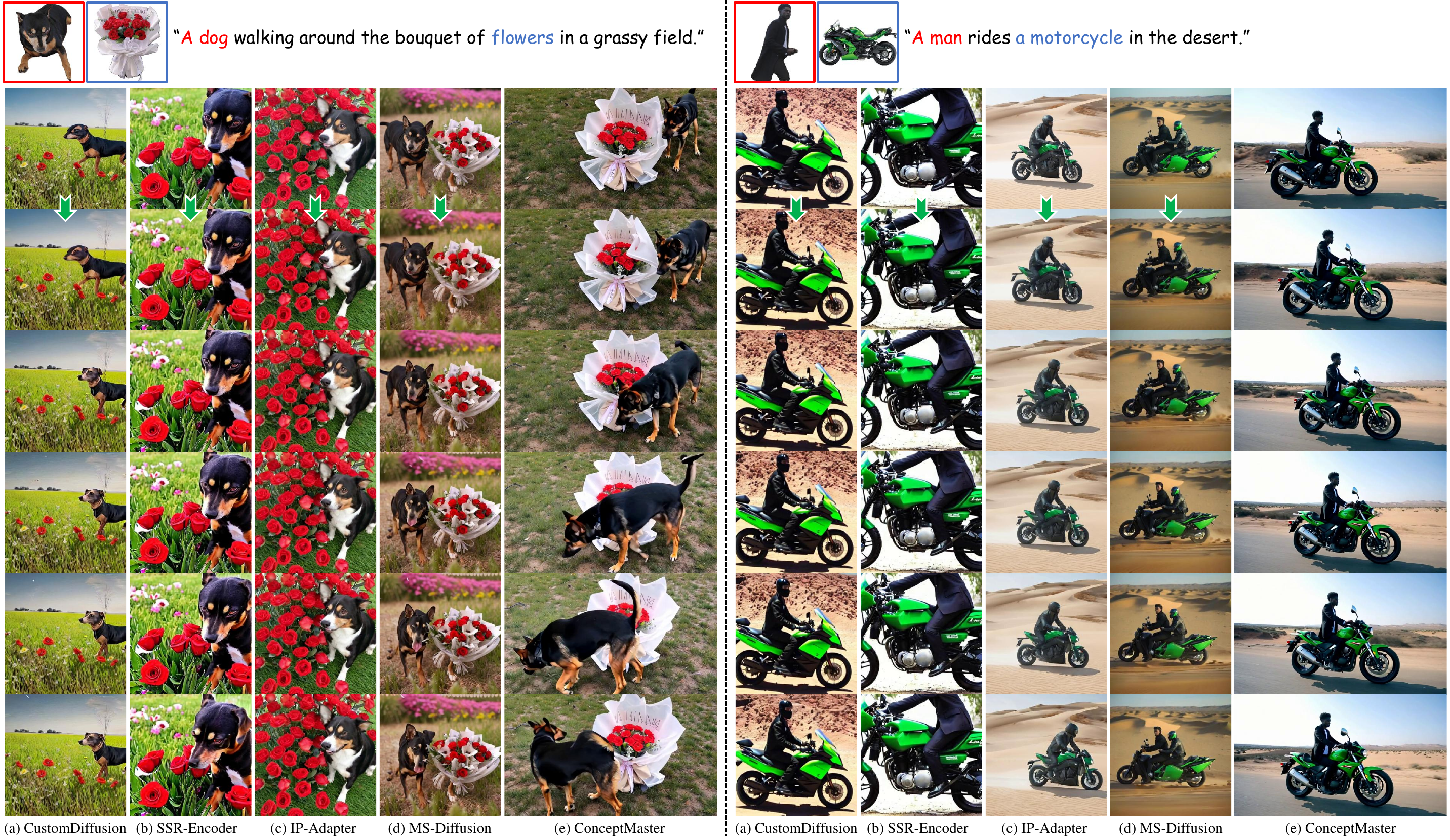}
\end{minipage}
\centering
\caption{More Qualitative comparison on multi-concept customization between ConceptMaster and naively combining the multi-concept image customization with image-to-video generation models.}
\label{fig:SuppComparison}
\vspace{-3cm}
\end{figure*}

We supplement the implementation details of the compared methods for the MCVC task.
We compare several open-sourced multi-concept image customization methods, including CustomDiffusion~\cite{kumari2023multi}, SSR-Encoder~\cite{zhang2024ssr}, IP-Adapter~\cite{ye2023ip} and MS-Diffusion~\cite{wang2024ms}, combining with the image-to-video (I2V) generation model I2VGen-XL~\cite{zhang2023i2vgen}, as a naive solution for the MCVC task with our ConceptMaster. 
While SSR-Encoder, IP-Adapter and MS-Diffusion do not need additional training and can be viewed as feed-forward customization models, CustomDiffusion requires users to manually few-shot examples for additional parameter training. 
Therefore, we use the collected reference images for each concept as training samples. Additionally, we follow its usage requirements, where we label specific character for each concept (\ie, \textit{s1* elephant is playing with s2* dog on the road}).
In addition to Tab.~\ref{table:DifferentMethods} and Fig.~\ref{fig:comparison} in the main paper, we demonstrate more quantitative results in Fig.~\ref{fig:SuppComparison}. 
Comparing with aforementioned methods, our end-to-end video customization models ConceptMaster could achieve better generation results more practicality than two-stage solutions.

%%%%%%%%%%%%%%%%%%%%%%%%%%%%%%%%%%%%%%%%%%%%%%%%%%%%%%%%%%%
% 5.
\section{More Discussions on Multi-Concept Embeddings Injection}

\begin{figure*}[t]
\centering
\small 
\begin{minipage}[t]{1.0\linewidth}
\centering
\includegraphics[width=1\columnwidth]{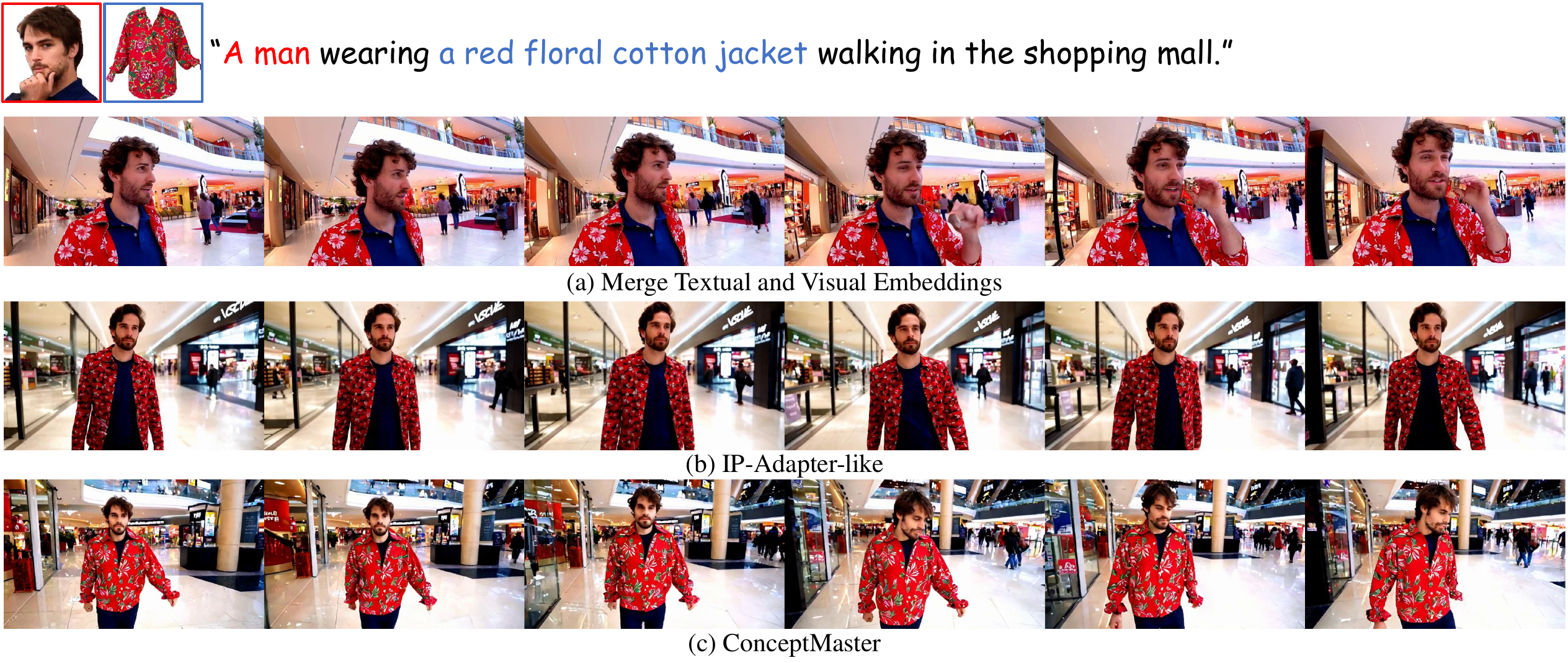}
\end{minipage}
\centering
\caption{More Qualitative comparison on different injection methods of multi-concept references.}
\label{fig:SuppTechnique}
\end{figure*}

We demonstrate more quantitative results between these three different multi-concept embeddings injection methods in Fig.~\ref{fig:SuppTechnique}.
Our key insight is to inject the represented multi-concept embeddings into the diffusion models in a standalone cross-attention layer.
While previous methods that the most representative ones include 1) BLIP-Diffusion~\cite{li2024blip}, which combines visual and textual caption embeddings as the whole condition representation. 2) IP-Adapter~\cite{ye2023ip}, which encodes the whole image as visual embeddings and aggregates into models by a decoupled cross-attention layer.
These two technical thoughts are widely adopted in previous image and video concept customization, which are different from our core insight.
% 这里修改
In addition to Tab.~\ref{table:DifferentMethods} and Fig.~\ref{fig:comparison} in the main paper, we demonstrate more quantitative results in Fig.~\ref{fig:SuppTechnique}, where both these two methods can hardly deal with the identity decoupling problem when there are similar concepts (\eg, a man and a girl). Even when concepts have huge semantic differences (\ie, a man and a red jacket), both two methods cannot maintain the concept fidelity of each concept.
Additionally, when merging the visual and textual embeddings together as the conditions, the instruction following ability becomes unsatisfactory and the dynamic degree also drops.
Therefore, our ConceptMaster adopts the most suitable manner of the injection of the multi-concept embeddings, which could represent and decouple multiple identities well.

%%%%%%%%%%%%%%%%%%%%%%%%%%%%%%%%%%%%%%%%%%%%%%%%%%%%%%%%%%%
% 6.
\section{More Discussions on Ablation Study}

\begin{figure*}[t]
\centering
\small 
\begin{minipage}[t]{1.0\linewidth}
\centering
\includegraphics[width=1\columnwidth]{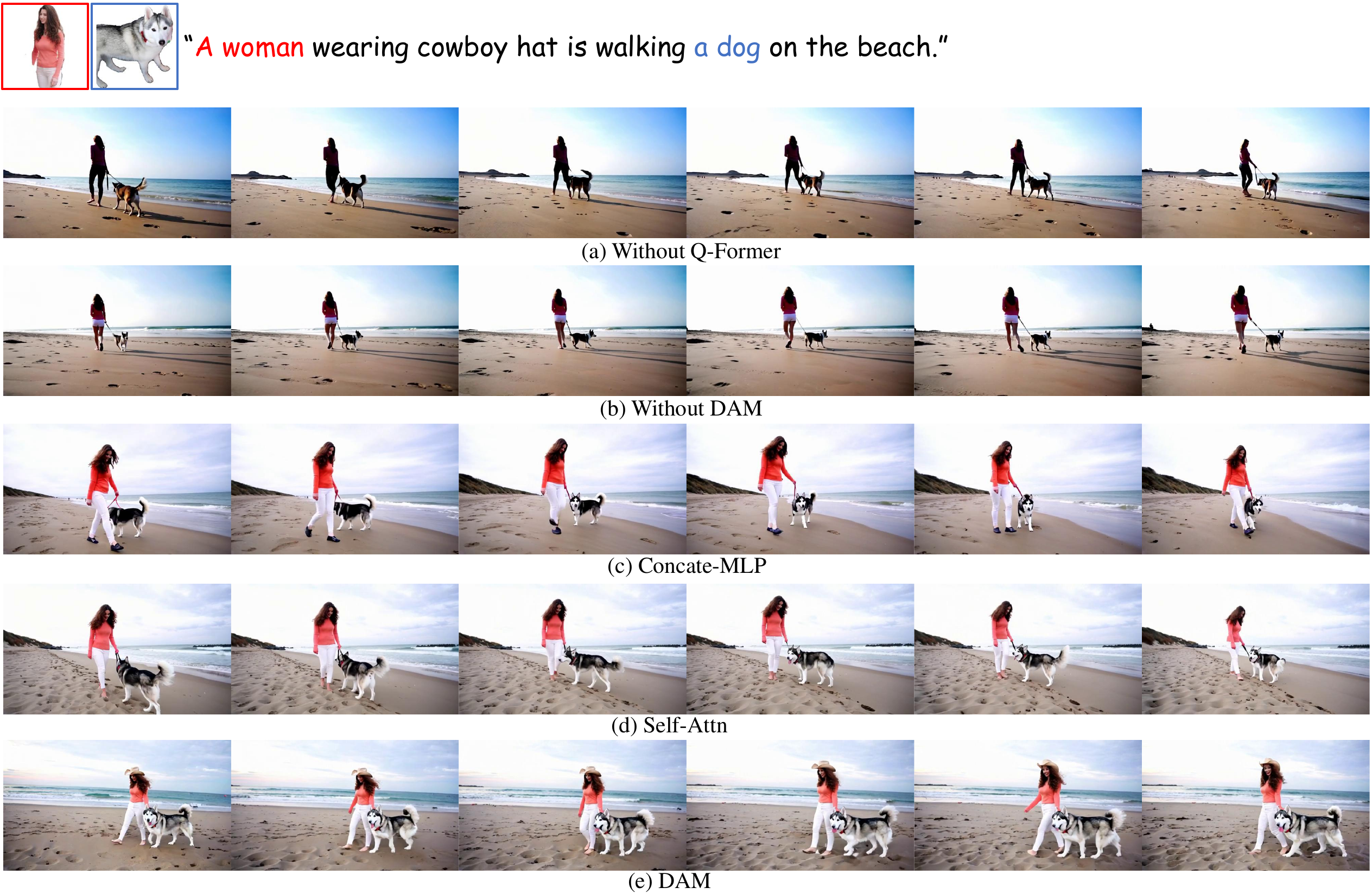}
\end{minipage}
\centering
\caption{More Qualitative comparison the effectiveness of the Q-Former and DAM modules.}
\label{fig:SuppDAM}
\end{figure*}

We demonstrate more quantitative results of the effectiveness of the Q-Former and DAM modules in Fig.~\ref{fig:SuppDAM}. 
% Q-Former
Initially, our ConceptMaster proposes to firstly utilizes a Q-Former network to integrate the dense visual tokens extracted by CLIP image encoder into the comprehensive visual embeddings.
% DAM
Additionally, we introduce the DAM module, which conducts the intra-pair attention module based on paired visual embeddings and textual descriptions representation.
% ablation
These designs demonstrate effective capability of decoupling and representing of multiple given references. While we also conduct several ablation architectures:
1) Replacing the Q-Former by an MLP layer, in Fig.~\ref{fig:SuppDAM} we can see that the visual appearances of provided \textit{woman} and \textit{dog} could no longer be captured by the generation results. Therefore, the Q-Former is significant to assist the representation of comprehensive visual embeddings.
2) Removing the DAM module, where only the extracted visual embeddings are further injected into the diffusion models, and the textual descriptions are unused. Similar phenomenon could be observed in Fig.~\ref{fig:SuppDAM} as the concepts could not keep their fidelity. This is because the absence of adequate text label representation fails to effectively represent and differentiate the uniqueness of multiple concepts, and it also results in poor alignment with the original diffusion space.
3) Replacing DAM by firstly concatenating the visual and textual embeddings along channel dimension, and downsampling the dimension to the original one by an MLP layer. While the insufficient representation would lead to the inharmonious combination of multiple concepts. Since when concatenating along the channel dimension and then downsamling by MLP, the tokens could not conduct interaction among them. As in Fig.~\ref{fig:SuppDAM}, the way that the woman who walks dog on the beach is unreasonable.
4) Replacing the intra-pair cross-attention by self-attention, where we firstly add the visual and textual embeddings and then conduct the self-attention operation. According to the qualitative and quantitative results in Tab.~\ref{table:Ablation} and Fig.~\ref{fig:ablation} in main paper and Fig.~\ref{fig:SuppDAM}, the cross-attention operation is better than self-attention, as the latter easily leads to more artifacts and inharmonious movements. 
Therefore, our proposed Q-Former and DAM modules would be the best designated architectures to simultaneously represent and decouple multiple references, and could create high-quality customized videos.

%%%%%%%%%%%%%%%%%%%%%%%%%%%%%%%%%%%%%%%%%%%%%%%%%%%%%%%%%%%
% 7.
\section{More Qualitative Results Demonstration}
\begin{figure*}[t]
\centering
\small 
\begin{minipage}[t]{1.0\linewidth}
\centering
\includegraphics[width=1\columnwidth]{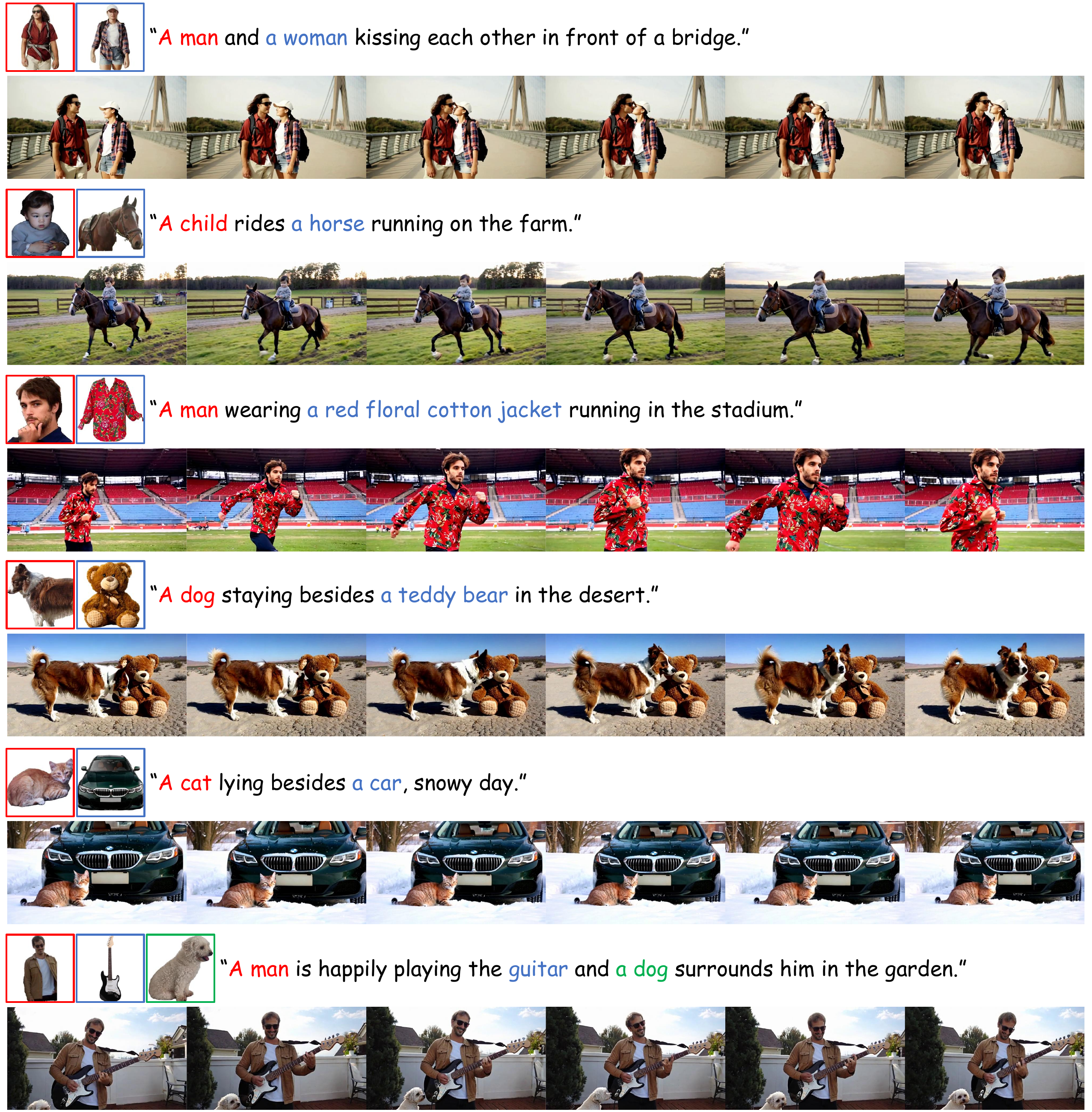}
\end{minipage}
\centering
\caption{More qualitative results of ConceptMaster on diverse scenarios (1/2).}
\label{fig:More1}
\end{figure*}

\begin{figure*}[t]
\centering
\small 
\begin{minipage}[t]{1.0\linewidth}
\centering
\includegraphics[width=1\columnwidth]{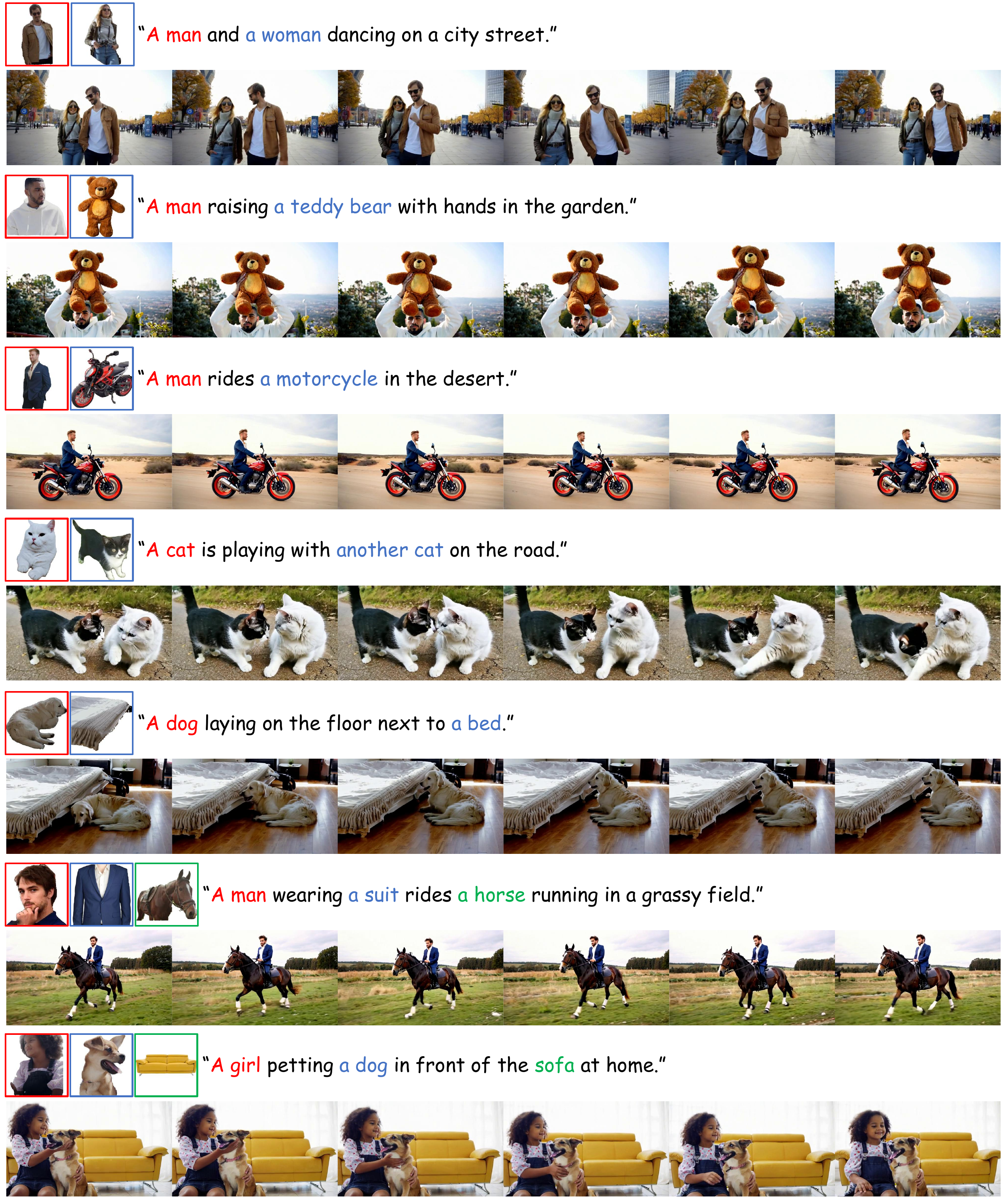}
\end{minipage}
\centering
\caption{More qualitative results of ConceptMaster on diverse scenarios (2/2).}
\label{fig:More2}
\end{figure*}

Our ConceptMaster could create high-quality and concept-consistent customized videos based on given multiple reference images in diverse scenarios, including but not limited to \textit{1) multiple persons, 2) persons with livings, 3) persons with stuffs, 4) multiple livings, 5) livings with stuffs and 6) persons with both livings and stuffs.}
We demonstrate more qualitative results including these scenes in Fig.~\ref{fig:More1} and Fig.~\ref{fig:More2}.

% MCBench prompt template
\clearpage
\begin{table}[h]
    \centering
    \begin{tabular}{|>{\raggedright\arraybackslash}m{4cm}|>{\raggedright\arraybackslash}m{12cm}|}
        \hline
        \textbf{Diverse Scenarios} & \textbf{Caption Templates} \\ \hline
        1) Multiple Persons & \textless ID1\textgreater\ and \textless ID2\textgreater\ hugging each other in front of a bridge. \\
          & \textless ID1\textgreater\ and \textless ID2\textgreater\ kissing each other in front of a bridge. \\
          & \textless ID1\textgreater\ and \textless ID2\textgreater\ walking down a city street. \\
          & \textless ID1\textgreater\ and \textless ID2\textgreater\ dancing on a city street. \\
          & \textless ID1\textgreater\ and \textless ID2\textgreater\ smiling and shaking hands in the office. \\
          & \textless ID1\textgreater\ and \textless ID2\textgreater\ walking on the beach. \\ \hline
        2) Persons with Livings & \textless ID1\textgreater\ walking \textless ID2\textgreater\ on the beach. \\
          & \textless ID1\textgreater\ walking \textless ID2\textgreater\ in the woods. \\
          & \textless ID1\textgreater\ petting \textless ID2\textgreater\ in the park. \\
          & \textless ID1\textgreater\ feeding \textless ID2\textgreater\ in the garden. \\
          & \textless ID1\textgreater\ and \textless ID2\textgreater\ running on the grass. \\
          & \textless ID1\textgreater\ rides \textless ID2\textgreater\ running on the farm. \\
          & \textless ID1\textgreater\ petting \textless ID2\textgreater\ in the stable. \\
          & \textless ID1\textgreater\ raising \textless ID2\textgreater\ in the garden. \\
          & \textless ID1\textgreater\ holding \textless ID2\textgreater\ and walking on the street. \\
          & \textless ID1\textgreater\ feeding \textless ID2\textgreater\ on the street. \\
          & \textless ID1\textgreater\ and \textless ID2\textgreater\ walking in the desert. \\ \hline
        3) Persons with Stuffs & \textless ID1\textgreater\ wearing \textless ID2\textgreater\ walking in the shopping mall. \\
          & \textless ID1\textgreater\ wearing \textless ID2\textgreater\ walking along the river. \\
          & \textless ID1\textgreater\ wearing \textless ID2\textgreater\ running in the stadium. \\
          & \textless ID1\textgreater\ wearing \textless ID2\textgreater\ dancing on the floor. \\
          & \textless ID1\textgreater\ rides \textless ID2\textgreater\ in the desert. \\
          & \textless ID1\textgreater\ rides \textless ID2\textgreater\ on the road. \\
          & \textless ID1\textgreater\ is happily playing \textless ID2\textgreater\ on the bench. \\
          & \textless ID1\textgreater\ is happily playing \textless ID2\textgreater\ in the desert. \\
          & \textless ID1\textgreater\ taking \textless ID2\textgreater, snowy winter. \\
          & \textless ID1\textgreater\ holding \textless ID2\textgreater\ and walking in the park. \\
          & \textless ID1\textgreater\ raising \textless ID2\textgreater\ in the garden. \\
          & \textless ID1\textgreater\ holding \textless ID2\textgreater\ and walking on the street. \\
          & \textless ID1\textgreater\ walking around \textless ID2\textgreater\ on the street. \\
          & \textless ID1\textgreater\ dancing in front of \textless ID2\textgreater, snowy day. \\
          & \textless ID1\textgreater\ walking in front of \textless ID2\textgreater\ on the street. \\
          & \textless ID1\textgreater\ dancing in front of \textless ID2\textgreater\ on the street. \\ \hline
        4) Multiple Livings & \textless ID1\textgreater\ is playing with \textless ID2\textgreater\ on the road. \\
          & \textless ID1\textgreater\ and \textless ID2\textgreater\ walking on the grass. \\
          & \textless ID1\textgreater\ walking around \textless ID2\textgreater\ in the desert. \\
          & \textless ID1\textgreater\ is playing with \textless ID2\textgreater\ on the street. \\ \hline
        5) Livings with Stuffs & \textless ID1\textgreater\ walking around \textless ID2\textgreater\ in a grassy field. \\
          & \textless ID1\textgreater\ playing with \textless ID2\textgreater\ in a grassy field. \\
          & \textless ID1\textgreater\ walking around \textless ID2\textgreater\ at home. \\
          & \textless ID1\textgreater\ is playing with \textless ID2\textgreater\ at home. \\
          & \textless ID1\textgreater\ is walking around \textless ID2\textgreater\ on the road. \\
          & \textless ID1\textgreater\ is staying besides \textless ID2\textgreater\ in the snow. \\
          & \textless ID1\textgreater\ is drinking water from \textless ID2\textgreater\ in the garden. \\
          & \textless ID1\textgreater\ is drinking water from \textless ID2\textgreater\ on the road. \\ \hline
        6) Persons with both Livings and Stuffs & \textless ID1\textgreater\ walking \textless ID2\textgreater\ in front of \textless ID3\textgreater\ on the street. \\
          & \textless ID1\textgreater\ and \textless ID2\textgreater\ walking around \textless ID3\textgreater\ in the snow. \\
          & \textless ID1\textgreater\ and \textless ID2\textgreater\ walking around \textless ID3\textgreater\ on the road. \\
          & \textless ID1\textgreater\ is happily playing \textless ID2\textgreater\ and \textless ID3\textgreater\ surrounds in the garden. \\
          & \textless ID1\textgreater\ taking \textless ID2\textgreater\ and walking with \textless ID3\textgreater\ on the grass. \\ \hline
    \end{tabular}
    \caption{Caption templates for Multi-Concept Evaluation Set.}
    \label{table:MCBench}
\end{table}

% S2.

% S4.

% S5.

% M1.

% M2.


\end{document}